\documentclass{IEEEtran}
\usepackage{cite}
\usepackage{amsmath,amssymb,amsfonts}
\usepackage{algorithmic}
\usepackage{graphicx}
\usepackage{textcomp}
\usepackage{enumitem}
\usepackage{soul}
\usepackage{amsmath}
\usepackage{txfonts}

\begin{document}

\title{Deep Convolutional Neural Network Ensembles Using ECOC}

\author{
  \IEEEauthorblockN{
   Sara Atito Ali Ahmed\IEEEauthorrefmark{1}, 
    Cemre Zor\IEEEauthorrefmark{3}, 
    Muhammad Awais\IEEEauthorrefmark{2}, 
    Berrin Yanikoglu\IEEEauthorrefmark{1} and 
    Josef Kittler\IEEEauthorrefmark{2}}
    \newline
    \\
  \IEEEauthorblockA{\IEEEauthorrefmark{1}Faculty of Engineering and Natural Sciences, Sabanci University, Istanbul, Turkey
      } \\
    \IEEEauthorblockA{\IEEEauthorrefmark{2}Centre for Vision, Speech and Signal Processing (CVSSP), University of Surrey, Guildford, United Kingdom} \\
    \IEEEauthorblockA{\IEEEauthorrefmark{3}Centre for Medical Image Computing (CMIC), University College London, United Kingdom}}

\maketitle

\begin{abstract}
Deep neural networks have enhanced the performance of decision making systems in many applications including image understanding, and further gains can be achieved by constructing ensembles. 
However, designing an ensemble of deep networks is often not very beneficial since the time needed to train the networks is very high or the performance gain obtained is not very significant. In this paper, we analyse error correcting output coding (ECOC) framework to be used as an ensemble technique for deep networks and propose different design strategies to address the accuracy-complexity trade-off. 
We carry out an extensive comparative study between the introduced ECOC designs and the state-of-the-art ensemble techniques such as ensemble averaging and gradient boosting decision trees. Furthermore, we propose a combinatory technique which is shown to achieve the highest classification performance amongst all.

\end{abstract}


\section{Introduction}\label{intro}

The improved generalisation capability of ensembles, compared to their constituent classifiers (also known as the \textit{base classifiers}), has long been established theoretically and experimentally. This capability is attributed to the complementarity of the individual classifiers in an ensemble, which jointly offer an error correcting mechanism, and is manifested in low prediction bias and variance  \cite{dietterich2000ensemble,kuncheva2003measures,zhou2012ensemble,polikar2012ensemble}. 
The combination rules employed by these systems can be as simple as taking a vote between the base classifiers, or more complex, where the classifiers are trained to compensate for the weaknesses of each other. 

Several fusion techniques such as averaging, majority voting \cite{penrose1946elementary}, bagging \cite{breiman1996bagging}, stacking \cite{dvzeroski2004combining}, random forests \cite{breiman2001random}, error correcting output coding \cite{kong1995error,guruswami1999multiclass} and their variants have been widely used in traditional machine learning. However, extensions of some of these approaches to deep learning (DL) systems 
have been deemed inefficient and challenging, due to the computational complexity associated with the training of deep networks, as well as the difficulty in securing diversity among the base classifiers. 
Therefore, most of the state-of-art DL ensembles are either formed of simple averaging (or voting) frameworks comprising only a small number of 
base classifiers \cite{deepface2014,szegedy2015going,redmon2016you,xiao2018deep,gessert2020skin}, or weak decision tree ensembles based on boosting deep features that have been already extracted \cite{badjatiya2017deep,ren2017novel,pang2019novel,ju2019model}.

Averaging  ensembles  are composed of base classifiers which are mainly obtained by modifying various DL elements such as the types of network architectures, and their parameters, data augmentation techniques, and learning meta parameters. 
An example is the DeepFace \cite{deepface2014}, where Taigman et al. construct a face verification system of 7 deep networks and achieve 97.35\% accuracy, compared to 97.0\% obtained using a single face verification network. 
In another work, Szegedy et al. \cite{szegedy2015going} increase the accuracy from 40\% (single network) to 43.9\% by averaging 6 GoogLeNet networks in the ILSVRC 2015 detection challenge. 
Yet another example is the winner of the PlantCLEF2017 competition \cite{lasseck2017image}, which is formed of 12 networks that are trained with an emphasis on complementarity and achieved a top-1 accuracy of 88.5\% in classifying 10,000 different plant species. 
Similarly, Gessert et. al. in \cite{gessert2020skin} employ multi-resolution EfficientNets \cite{tan2019efficientnet} for skin lesion classification based on an ensemble of 15 deep networks, where the area under the curve (AUC) is increased from an average of $94$ per classifier to $95.4$.  

Despite the performance gain achieved by the deep averaging ensembles, the increased time complexity, which scales linearly with the addition of each base classifier network, comes out as the main drawback. 
In the literature, gradient boosting decision tree (GBDT) methods are proposed to address this shortcoming, by operating on the deep features obtained from one base network (contrary to generating many deep networks as base classifiers) and constructing a sequential ensemble of trees which are trained to correct each other's errors, using these features. 

There are three commonly used GBDT variations in the literature: extreme gradient boosting (XGBoost) \cite{chen2016xgboost}, light gradient boosting machine (LightGBM) \cite{ke2017lightgbm}, and categorical  boosting  (CatBoost) \cite{prokhorenkova2018catboost}. As an example of XGBoost, Pang et al. \cite{pang2019novel} propose a subcellular localisation method by integrating the Convolutional Neural Network (CNN) and XGBoost, where CNN acts as a feature extractor and XGBoost acts as a classifier to identify the protein subcellular localisation. In another literature review, Torres-Barr{\'a}n et al. \cite{torres2019regression} study the application of XGBoost to global and local wind energy prediction and solar radiation problem, exploiting gradient boosting regression methods. As for LightGBM, Ju et al. in \cite{ju2019model} overcome the limitation of the single-convolution model in predicting the wind power by integrating the LightGBM 
algorithm to improve the robustness and accuracy of the forecasting.

Although GBDT is a powerful ensemble technique, the major disadvantages are its inability to deal with a high number of classes and the high number of hyper-parameters that need to be tuned to obtain the desired performance. It is important to note that the improved time complexity obtained with respect to the averaging ensembles is at the expense of a reduced ensemble performance. In this article, we address the drawbacks of deep averaging ensembles (time complexity) and GBDT (accuracy), and propose an efficient DL framework based on error correcting output coding (ECOC).

ECOC, borrowed originally from the communication theory \cite{hocquenghem1959codes,bose1960class}, is a multi-class classification ensemble, in which a given multi-class problem is decomposed into several two-class problems, whose simpler decision boundaries are then combined to give the final, more complex decision boundary. The errors of the base classifiers that implement the two-class decision boundaries are corrected to a certain degree \cite{Dietterich95}. 
Several data-dependent and independent approaches can be used for guiding the decomposition process \cite{pujol2006discriminant,bautista2012}.
In \cite{james1998error}, it has been theoretically and experimentally demonstrated that ECOC frameworks formed using random class splits obtain close to Bayes performance, if there are infinitely many such splits, and the associated base classifiers achieve good accuracy. In practice, the performance reaches the optimum very rapidly \cite{james1998majority} as the number of the classifiers increases. The superiority of this method, which we refer to as \textit{randECOC}, over the rest of the data independent and dependent ECOC approaches is demonstrated in \cite{Dietterich95,james1998error,EscaleraDecoding}.

Although ECOC has been commonly employed in traditional machine learning applications \cite{xiao2010speech,ye2011pedestrian,smith2015facial,gu2019active}, to date, to the best of our knowledge, its potential as a method of constructing deep convolutional neural network ensembles has neither been exploited nor analysed.
The only work addressing ECOC in DL research is \cite{zhang2020adversarial}, where it is utilised for the adversarial robustness of the networks. 

In this work, by operating on the base network features, we propose and analyse efficient implementation strategies for randECOCs. We investigate three different design procedures: i) 
the straightforward approach of training the base classifiers independently,
ii) multi-task learning (MTL) for faster training, and iii) MTL with embedded error correction.
It is expected that the selection of the most appropriate design procedure will be carried out by the user, depending on the specific requirements of an application, and the time complexity versus accuracy trade-off. 

The systems are evaluated on four public datasets: CIFAR-10 and CIFAR-100 (10 and 100 classes, respectively) for object recognition \cite{krizhevsky2009learning}, and SVHN of Google street view images of house numbers (10 classes) \cite{netzer2011reading}, and PlantVillage dataset \cite{hughes2015open} consisting of $38$ plant leaf disease types.

We show that for all the proposed design techniques, randECOC almost always surpasses the GBDT performance at comparable time complexity, 
when MTL based implementation strategies are considered. 
When compared with averaging ensembles, a degradation in performance has been noted, due to the end-to-end training nature of averaging ensembles as opposed to the feature-based training of randECOC. However, for the users who have enough resources to accommodate averaging ensembles, we propose combining randECOC with averaging, and show that this setup guarantees the best performance with the highest accuracy in all scenarios. 

The main contributions of this study can be summarised as follows:
\begin{itemize}

   
    \item We propose three different designs for randECOC ensembles to be used with convolutional deep neural networks and analyze these approaches in terms of accuracy and time complexity trade-off, using several different deep network architectures and 4 different datasets.
    
         \item We perform an empirical comparison of the randECOC ensembles and state-of-the-art ensemble methods for deep learning, i.e. ensemble averaging and GBDTs,
         and show that the proposed MTL strategies provide the best time complexity versus accuracy trade-off.
     
    \item We propose a hybrid approach, combining randECOC strategies and  ensemble averaging, to achieve state-of-the art classification performance for all network and dataset combinations.
    
\end{itemize}

The article is structured as follows.  Section \ref{background} provides  a background information on the state-of-the-art deep ensemble classification techniques as well as the ECOC framework. In Section \ref{method}, different ECOC training strategies using features extracted by deep convolutional neural networks are presented. This is followed by  their experimental analysis in Section \ref{exp} and a discussion of the results obtained. Finally, conclusions of this study are presented in Section \ref{conc}.

\section{Background}\label{background}

In the literature, averaging and majority voting are the most commonly used classifier combination approaches where the ensemble output is calculated based on the (weighted) average of the base classifier outputs or their mostly voted prediction. Bagging \cite{breiman1996bagging} is a special case of majority voting for which the base classifiers are trained on differently versions \cite{efron1992bootstrap} of the same data obtained by resampling. More complex combination rules include methods such as boosting \cite{schapire2003boosting}, where the classifiers are trained sequentially to compensate for the weaknesses of those already selected and stacking \cite{wolpert1992stacked}, where the outputs of all classifiers are fed into a new model to generate the final prediction. Another commonly used ensemble technique is random forests \cite{breiman2001random}, which are composed of multiple decision trees trained on  bootstrapped training data with an additional step of  feature-bootstrapping to allow for a random selection (with replacement) of features at each tree node. The final decision is based on the (weighted) average of the outputs or the majority vote of the individual tree decisions. 

The fusion rules most commonly applied in the state-of-the-art deep learning ensembles are based on averaging or majority voting. These ensembles consist of a small number of deep neural network architectures as base classifiers, which differ from each other in terms of the data augmentation techniques used during training and / or network architectures and / or learning parameters (such as learning rates, training and validation set partitions, weights initialisation and data batches). Due to the costly training of these ensembles, they typically are composed of only a handful of base classifiers. 

Overcoming the time complexity of the averaging / voting ensembles of deep neural networks, the second most common combination strategy, 
gradient boosting decision trees (GBDT), depends on extracting the bottleneck features of one \textit{base network} and using them for training a sequence of decision trees. However, the gain in time complexity of this approach is compromised by  reduced accuracy, especially for high number of classes. Moreover, the method requires a high number of hyper-parameters to be tuned to obtain the desired performance.
 
In Section \ref{GBDTs}, we analyse three state-of-the-art variants of the GBDT method found in the literature; namely,  extreme gradient boosting (XGBoost) \cite{chen2016xgboost}, light gradient boosting machine (LightGBM) \cite{ke2017lightgbm} and categorical boosting (CatBoost) \cite{prokhorenkova2018catboost}, in detail. Later on in Section \ref{ECOC}, we provide the background for error correcting output coding (ECOC) ensembles,  on which we  build our novel design strategies for designed ensembles of deep learning networks, presented in Section \ref{method}.

\subsection{Gradient Boosting Decision Trees (GBDT)} \label{GBDTs}
Gradient boosting is a machine learning technique for regression and classification problems that creates an ensemble of weak prediction models to achieve powerful prediction. When decision trees are used as the base classifiers, the method is referred to as gradient boosting decision trees (GBDT).

Unlike random forests, where the decision trees are constructed in parallel prior to combination, GBDT employs a boosting approach,
in which each tree is sequentially trained with the aim of correcting the error produced by its predecessor. In particular, every tree is trained to learn the residual between the desired output and the output of the previous tree, using gradient descent.The most important parameter in GBDT is the number of base classifiers which controls the model complexity. The most recent and efficient GBDT methods developed are XGBoost \cite{chen2016xgboost}, LightGBM \cite{ke2017lightgbm}, and CatBoost \cite{prokhorenkova2018catboost}. These algorithms differ from each other in terms of the  mechanism used for splitting the tree nodes.

Extreme gradient boosting (XGBoost) \cite{chen2016xgboost}, is a highly extensible tool mainly designed to overcome the overfitting limitations of the traditional gradient boosting methods. It uses pre-sorted and histogram-based algorithms for computing the best split, which continues until the maximum level, pre-defined by the ``max\_depth'' hyper-parameter, is reached. Once at the maximum level, the splits are pruned backwards until there is no positive gain.

Light gradient boosting machine (LightGBM), proposed and developed by Microsoft \cite{ke2017lightgbm}, uses gradient-based one-side sampling (GOSS) to filter out data instances on the basis of their contribution to the  gradient of the loss function. The best split is obtained by using all of the instances with large gradients and a random sample of instances with small gradients to maintain a balance between the training data  reduction and  accuracy. LightGBM uses a leaf-wise tree growth mechanism which allows the growth of an imbalanced tree. 

Categorical boosting (CatBoost) \cite{prokhorenkova2018catboost} focuses on categorical features by using minimal variance sampling (MVS) which is a weighted sampling method at the tree-level. Unlike LightGBM, CatBoost grows balanced trees which makes this method less prone to overfitting, and uses combinations of categorical
features as additional categorical features  to capture high-order dependencies. As it is infeasible to process all of the possible combinations, CatBoost solves the exponential growth of the features combination by constructing the combinations in a greedy way.

\subsection{Error Correcting Output Coding (ECOC)} \label{ECOC}

Error Correcting Output Coding (ECOC) is a generic ensemble classification framework designed for multi-class classification problems \cite{Dietterich95}, where the aim is to decompose a given multi-class problem into several two-class problems. The final decision boundary is formed by combining the boundaries of the base classifiers trained on these simple decompositions while providing a scope for error correction. 

The way the decomposition is carried out in ECOC is defined by a \textit{design code matrix}. Accordingly, a base classifier may be assigned the task of separating a particular class from all of the others, or learning a random dichotomy of the classes. The commonly used ensemble approaches such as one-vs-one or one-vs-all can therefore be considered as special types of ECOC systems.

Let us  consider a problem with $K$ classes $\{c_1,c_2, \ldots c_K\}$, $L$ base classifiers $\{h_1, h_2, \ldots h_L\}$, and a pre-designed code matrix $\mathbf{M}$ of size $K \times L$ as illustrated in Table \eqref{ECOCSample}, for $K=4$ and $L=5$. A particular element $M_{ij} \in \{+1,-1\}$ indicates the desired label for class $c_i$ to be used in training the base classifier, $h_j$. For instance in Table \eqref{ECOCSample}, the base classifier, $h_1$, is assigned the task of separating instances belonging to classes $c_1$ and $c_2$ from instances belonging to classes $c_3$ and $c_4$. The classes $c_1$ and $c_2$ are re-labelled with label +1, while $c_3$ and $c_4$ are re-labelled with label -1, to reflect this two-class problem.

The design (encoding) of the code matrix can be carried out in several ways. These include problem-independent approaches such as one-vs-one or one-vs-all \cite{Dietterich95}, or problem-dependent methodologies where the aim is to split the classes in the given data domain \cite{EscaleraDecoding,lorena2010building} meaningfully.

In decision making (testing), firstly, a given test instance $\mathbf{x}$ is classified by each base classifier to obtain the output vector $\mathbf{Y} = [y_1, ..., y_L]$ where $y_j$ is the hard or soft output of the classifier $h_j$ for $\mathbf{x}$. 
Then, the distance between $\mathbf{Y}$ and the \textit{codeword} $\mathbf{M_i}$ of class $c_i, \forall i$, is computed using a metric such as Hamming, Manhattan or  Euclidean distance. 
The class $c*$ associated with the minimum distance is  chosen as the predicted class, such that 
\begin{equation}
c*=\arg\min_{i=1\ldots k}\text{dist}(\mathbf{Y},\mathbf{M_i})
\label{eq:ecocdec}
\end{equation}

While choosing the closest codeword during the target prediction, the system is able to correct some of the base classifiers mistakes. Specifically, up to $\left\lfloor (e-1)/2\right\rfloor$  base classifier errors can be corrected if Hamming Distance (HD) is chosen as the distance metric, and the minimum HD between any pair of codewords is $e$. 



\begin{table}[h]
\centering
\caption{A sample ECOC matrix for a 4-class
classification problem with 5 base classifiers}

\begin{tabular}{|c|c|c|c|c|c|} \hline
 & $h_1$   &  $h_2$   &  $h_3$   &  $h_4$   &  $h_5$  \tabularnewline\hline  
 $c_1$   &  +1   &  +1   &  +1   &  -1   &  -1   \tabularnewline \hline
 $c_2$   &  +1   &  -1   &  -1   & +1   &  -1   \tabularnewline \hline
 $c_3$   &  -1   &  +1   &  -1   & +1   &  -1   \tabularnewline \hline
 $c_4$   &  -1   &  -1   &  -1   & -1   &  +1   \tabularnewline \hline
\end{tabular}
\label{ECOCSample}
\end{table}

Although the encoding and decoding of ECOC matrices are open research problems, it is important to note that randomly generated ECOC matrices (randECOC) have been shown to reach Bayes performance when used with large enough number of base classifiers, each of which exhibiting close to Bayes accuracy \cite{james1998error}. In practice, it has been experimentally demonstrated in \cite{james1998majority} that for problems involving ${\sim}10$ classes, randECOCs of length 20-30 would be enough to converge to optimum performance, whereas this number would grow to 200-300, when the number of classes is ${\sim}100$.
\section{Design Strategies for \MakeLowercase{rand}ECOC Using CNNs}\label{method}

Under the assumption of unconstrained computational resources, the optimal strategy to achieve the highest prediction performance using randECOC would be to train each base classifier independently.
End-to-end training of these classifiers, each of which is initialised with random weights, would help increase the diversity between classifiers and enforce independence which is a key element in achieving close-to-Bayes performance \cite{james1998majority,james1998error}. However, this procedure would suffer from similar time complexity drawbacks as in averaging ensembles and be impractical in real-life applications.

For this reason, in this section, we propose and analyse different design
strategies for randECOC matrices, which address the shortcomings of time complexity associated with averaging ensembles, while still achieving better performance than their time efficient alternative, GBDT. In the design strategies presented in Section \ref{indepbase} through \ref{en2end_MTL}, we propose to initially train a multi-class \textit{base network} to obtain the bottleneck features (as opposed to end-to-end training), and build three implementation techniques with varying accuracy vs time complexity trade-off on these features.

Specifically, after presenting the straightforward approach to designing randECOC ensembles with base classifiers trained independently using bottleneck features in Section \ref{indepbase},
we propose a more time-efficient implementation strategy based on multi-task learning (MTL) in Section \ref{mtlone}. Then, in Section \ref{en2end_MTL}, the MTL based strategy is further improved with the incorporation of an error-correcting mechanism as a separate layer of the network. 
This strategy aims to couple the base classifier training to the classification problem, as opposed to training the base classifiers only to be in agreement with the encoding matrix:
Few research work exists to learn or modify the ECOC matrix after the training of the base classifiers, for their joint optimization  \cite{ethem99,flipiscis2010,beamECOC2016}.
For all three designs, the base networks are assumed to be  convolutional neural networks (CNNs). 

\subsection{Independent Learning of Base Classifiers}
\label{indepbase}

\begin{figure}[thb]
\begin{center}
\includegraphics[width=0.99\linewidth]{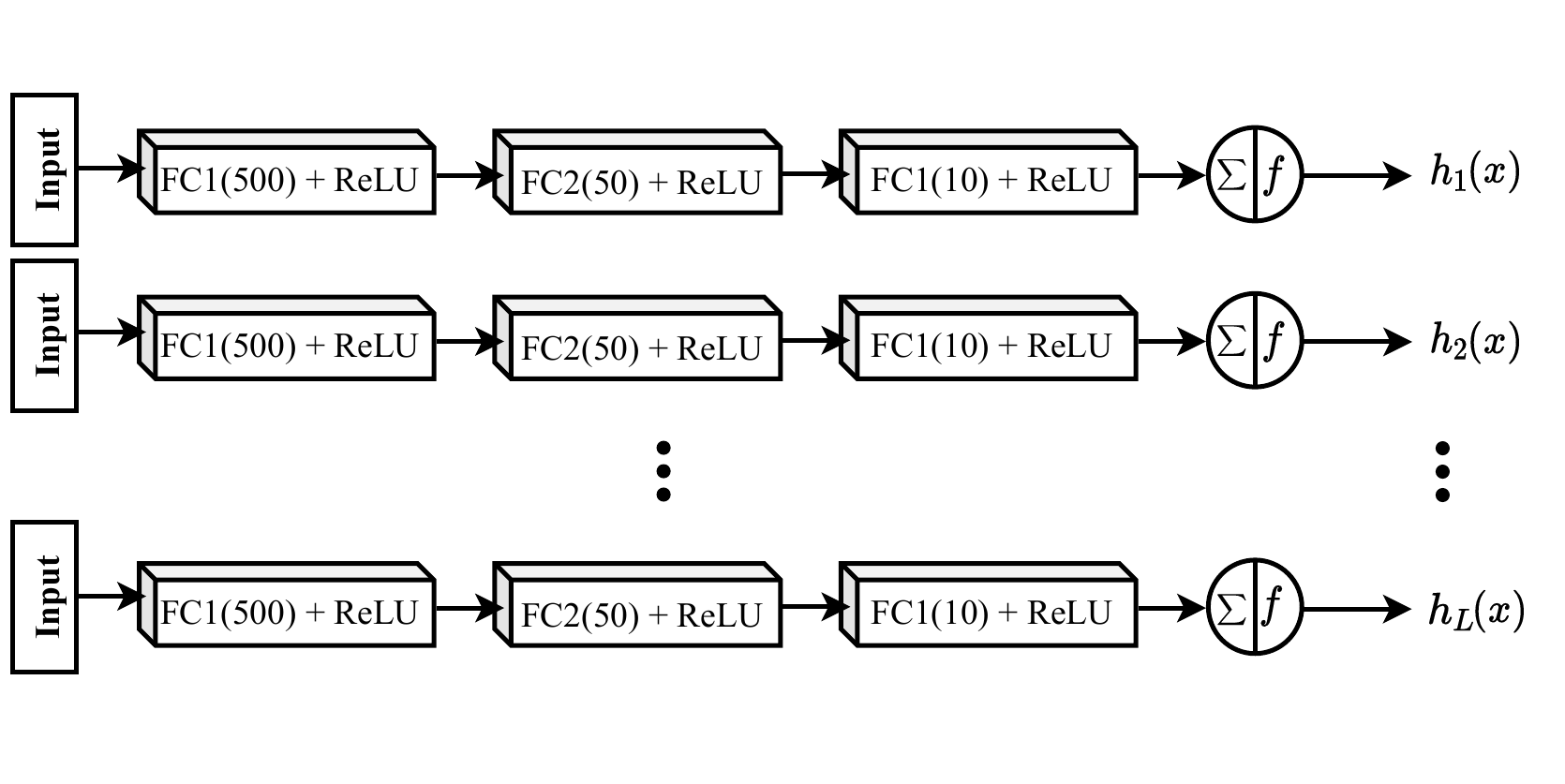}
\vspace{-0.4cm}
\caption{An independent base classifier architecture with a 3-hidden layer shallow network, consisting of fully connected layers followed by rectified linear units, one for each base classifier of the ECOC ensemble. The input comprises the features extracted by the bottleneck layer of a trained \textit{base network}.
}
\label{figindep}
\end{center}
\end{figure}

In this approach, the base classifiers are trained one by one and independently according to a given randECOC matrix, using the deep features extracted from the bottleneck layer of a base network. 
A schematic diagram illustrating an example of independently trained base classifier networks is given in Figure \ref{figindep}.

Here, we propose to design the base classifiers as shallow networks, whose outputs are then combined for an error-correcting randECOC decoding to give the final output.
In other words, after extracting the output vector $\mathbf{Y(x)}$ for  a given test sample $\mathbf{x}$ from all shallow networks, 
the prediction is carried out in a \textit{separate} decoding step, where $\mathbf{x}$ is assigned the class with the  closest codeword to $\mathbf{Y(x)}$ (see Equation \ref{eq:ecocdec}).

\subsection{Multi-task Learning of Base Classifiers}
\label{mtlone}

In order to achieve close-to-Bayes accuracy, the number of base classifiers required for a randECOC ensemble should increase with the number of classes. Although all independent tasks can potentially be trained in parallel as proposed in Section \ref{indepbase}, this framework might be  unattractive under the assumption of limited resources, despite the performance gain promised. 

To address this, we consider the idea of simultaneous training of the base classifiers by employing an MTL based strategy, where the classifiers are trained to learn multiple labels, i.e. the desired base classifier outputs, at the same time. Although this method can only approximate the performance of the independently trained base classifiers, it is important from the  point of view of accuracy versus time complexity trade-off.

\begin{figure}[th]
\begin{center}
\includegraphics[width=0.99\linewidth]{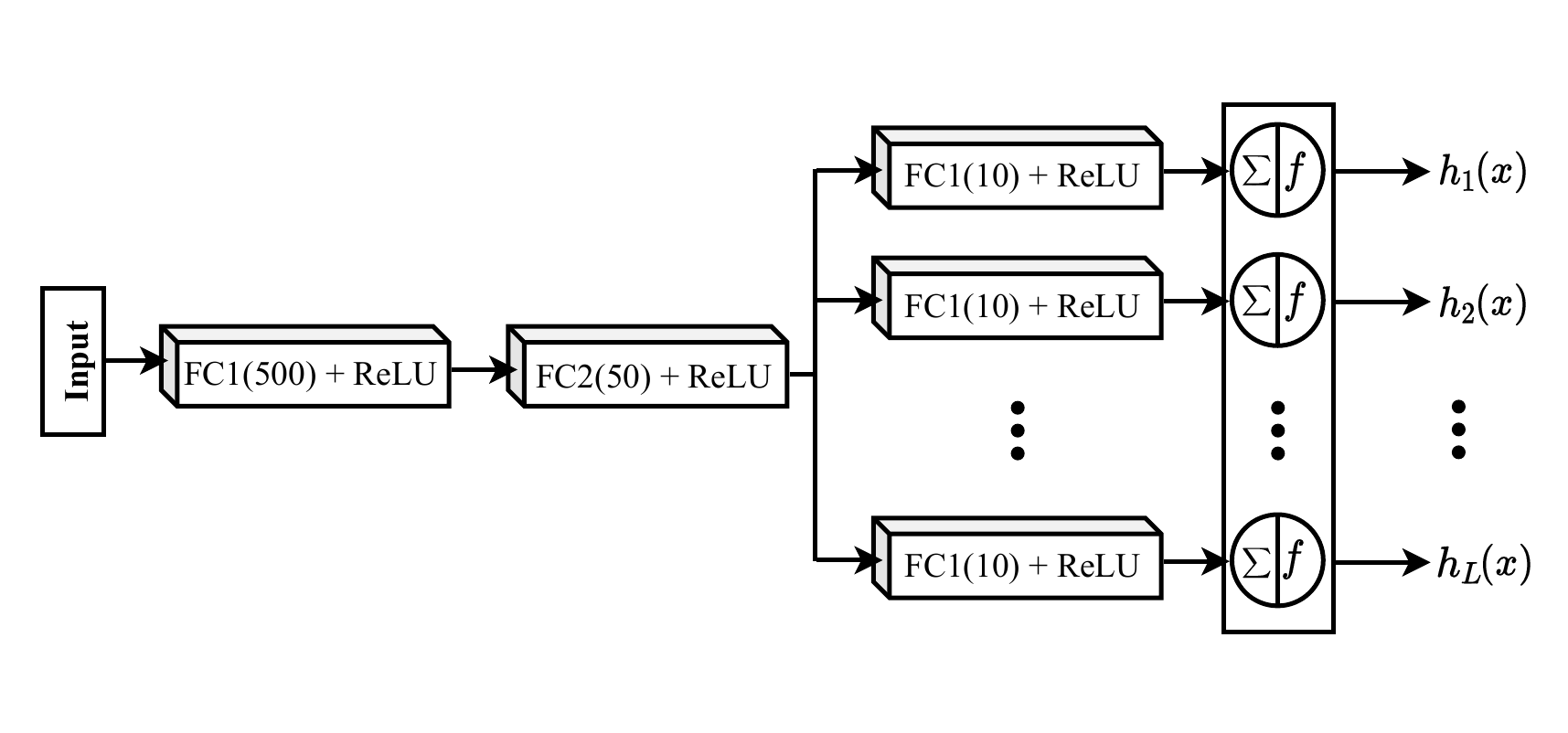}
\vspace{-0.5cm}
\caption{Multi-task learning architecture, with two shared modules and one classifier specific module. All layers are fully connected networks with rectified linear units.}
\label{MTLbasic} 
\end{center}
\end{figure}

In this approach, we have a single MTL network comprising several shared layers among all base classifiers, with $L$ output nodes; as opposed to $L$ independent networks.  In other words, while training the independent classifiers sequentially would mean the repetition of the randECOC procedure $L$ times, training all classifiers at the same time via MTL would imply carrying out this step only once. Hence, the time complexity of the MTL network is approximately $L$ times better than the independent sequential training. An illustration of an example MTL network is presented in Figure \ref{MTLbasic}.

The prediction is carried out in the same way as in Section \ref{indepbase}, where ECOC decoding is executed as the second step following the extraction of classifier outputs in the first step. Note that we propose this network should also include a small number of shallow, classifier specific layers to allow for diversity.

As a further advantage of the MTL network, it should be noted that the sharing of the base network and the subsequent layers are expected to reduce overfitting, as observed in literature \cite{aly2018multi}, since the nodes in the shared layers are constrained to work for multiple classifiers.

\begin{figure*}[thb]
\begin{center}
\includegraphics[width=0.8\linewidth]{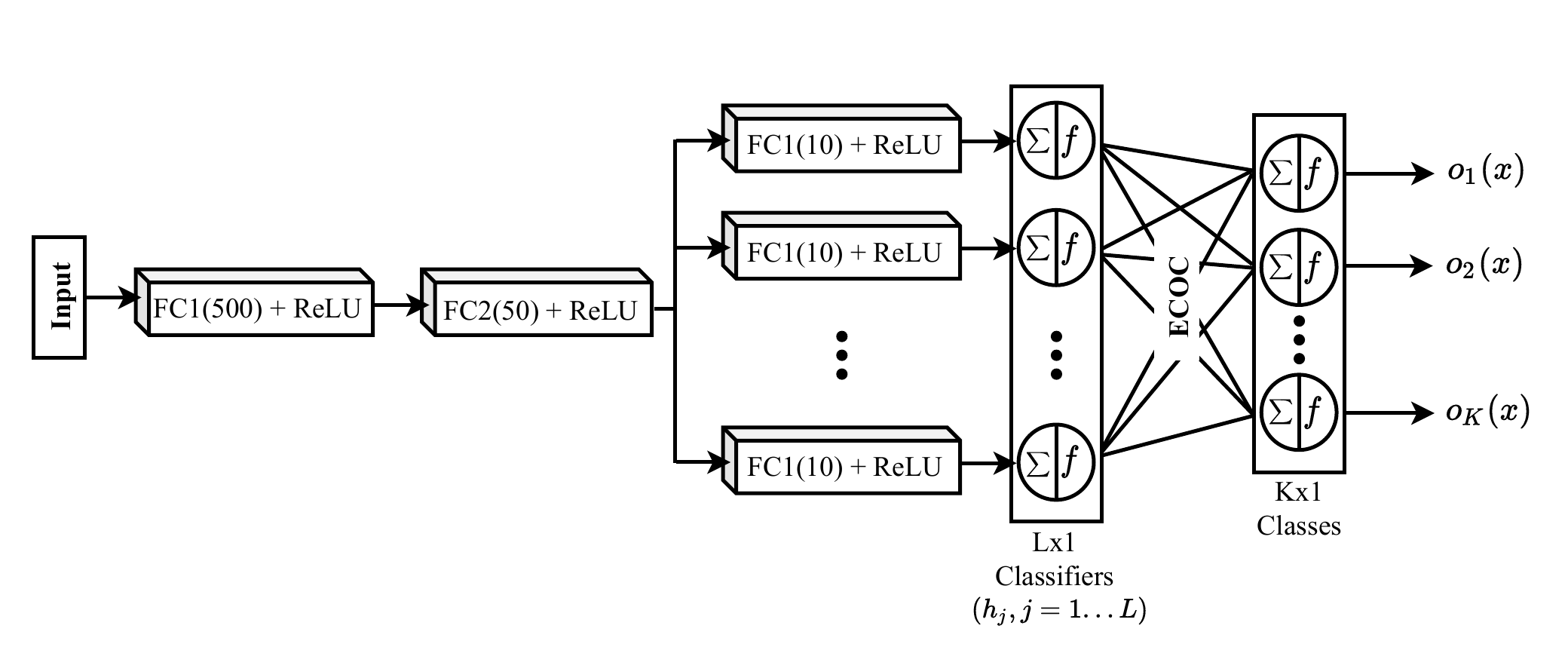}
\caption{Multi-label architecture with embedded ECOC decoding, including two shared modules and one classifier specific module. The base classifier output layer is followed by the ECOC embedding layer with fixed weights. The output $o_i$ corresponds to the score of class $c_i$.} 
\label{MTL-e2e}
\end{center}
\end{figure*}
\subsection{Multi-task Learning with Embedding}
\label{en2end_MTL}

Despite its advantages in terms of speed and reduced overfitting, the MTL network described in Section~\ref{mtlone} is suboptimal in the sense that the second step of the prediction, namely ECOC decoding, is carried out separately from the network training.
In other words, while the base classifiers are enforced to learn the dichotomies (two-class problems) indicated by the randECOC matrix, they are not enforced to reveal the desired \textit{multi-class} label. 

In order to address this issue, we propose to extend the MTL network with a K-node output layer, with weights set from the randECOC codewords and the output nodes representing the original classes. This layer not only enforces the final, multi-class decision on the outputs of the two-class base classifiers, but also includes the ECOC decoding inherently. The proposed framework is illustrated in Figure~\ref{MTL-e2e} with an example architecture. It is referred to as ``MTL w/ embedding" in the remainder of this paper.

It is worth mentioning that the randECOC matrix is not learned here but is pre-set. In some earlier work, the matrix was modified during or after the training of the base classifiers, with the  goal of reducing this decoupling between the encoding and base classifier training stages \cite{ethem99,beamECOC2016,flipiscis2010}.

Let us assume that the nodes corresponding to the base classifiers $h_j$, $j=1\ldots L$ 
are connected to the output nodes $o_i, i=1\ldots K$ with the preset ECOC matrix weights $w_{ij}=M_{ij}$. For a given input $\mathbf{x}$, each output node $o_i$ represents the 
score for class $c_i$, such that

\begin{equation} \label{oi}
o_i(\mathbf{x}) = \sum_{j=1}^{L}{h_j(\mathbf{x}) \times w_{ij}  }  = \mathbf{h(x)} \cdot \mathbf{w_i}.
\end{equation}
Note that the maximum value of  $o_i(\mathbf{x})$ is $L$ when all the base classifier outputs are in agreement with their associated bits of the codeword for that class (targets); 
while the minimum is $-L$ when all base classifier outputs are wrong. In other words,
\begin{equation}\label{Hdequal}
HD(\mathbf{w_c},\mathbf{h(x)})= \frac{L-o_c(\mathbf{x})}{2}.
\end{equation}

The loss function used to train the network is designed with two goals: 1) To maximise the output of the correct class, $o_c$; 
2) To match the output vector $\mathbf{h(x)}$ to the predetermined codeword $\mathbf{w_c}$, so as to benefit from the ECOC framework. 
Therefore, given a sample of class $c$ and groundtruth $\mathbf{T}=[t_1 \ldots t_K]$
(one-hot encoded vector where $t_{c}=1$ for only the correct class and zero elsewhere), we use the loss function given in Equation \ref{Eq:Loss}. 
We ignore $o_i, i \neq c$ because maximizing $o_c$ is equivalent to minimizing other class outputs, thanks to the design of the ECOC matrix. 

\begin{equation}\label{Eq:Loss}
L = (L - o_{c}(\mathbf{x}))^2  + \sum_{l=1}^{L}{(h_l(\mathbf{x}) - M(c,l))^2}
\end{equation}

With ternary ECOC where there are zeros in the code matrix, the maximum output value of $L$ is not attainable for $o_c$, hence $L$ should be replaced with the  number of non-zeros in a codeword.

{To train the network, we use stochastic backpropagation, starting with the weights of the base classifiers $h_j$, as the ECOC matrix weights are fixed. The partial derivative of our combined loss function with respect to $h_j(\mathbf{x})$ is computed as:}


\begin{align*} \label{Eq:partial}
\partial L/\partial h_{j}\left(x\right) & =\frac{\partial\left(L-o_{c}\left(x\right)\right)^{2}}{\partial o_{c}\left(x\right)}\frac{\partial o_{c}\left(x\right)}{\partial h_{j}\left(x\right)}+\sum_{l=1}^{L}\frac{\partial\left(h_{l}\left(x\right)-M\left(c,l\right)\right)^{2}}{\partial h_{j}\left(x\right)}\\
 & =2\left(L-o_{c}\left(x\right)\right)\omega_{cj}+2\left(h_{j}\left(x\right)-M\left(c,j\right)\right)
\end{align*}


For the final prediction, the class $c_i$ that has the 
maximum $o_i(\mathbf{x})$ (equivalently, minimum distance to the base classifier outputs $\mathbf{h(x)}$) is chosen as the correct class.

\section{Experimental Analysis and Results} \label{exp}

To evaluate the effectiveness of the proposed randECOC techniques and compare their efficiency in terms of time complexity and accuracy with the state-of-the-art ensemble methods, we 
conduct various experiments using well-known deep architectures and  multi-class datasets. 
Specifically, the comparative studies are performed on:

\begin{enumerate}
    \item Simple averaging ensemble;
    \item Gradient boosting decision trees (GBDTs): XGBoost, LightGBM, and CatBoost;
    \item randECOC ensembles: Independent learning, MTL,  and MTL with embedding.
\end{enumerate}


After carrying out the comparisons, we combine randECOC and GBDT approaches with ensemble averaging,
i.e. we generate \textit{ensembles} of randECOC and GBDT ensembles and analyse their performance.
The purpose of this experiment is to measure the  
highest possible prediction accuracy, 
for scenarios where the available resources (computational resources including processing power, time and storage) are not a limiting factor for the user.

In Section \ref{datasets}, the details of the datasets used in the experiments are presented and in Section \ref{basenetworks}, various base network architectures utilised in this study are described. This is followed by providing the details of the experimental setup in Section \ref{expset}, and the thorough discussion of the results in Section \ref{results}.

\subsection{datasets}
\label{datasets}
We carry out the experimental analyses on four
state-of-the-art multi-class classification problems based on digit classification and object recognition using images. In all tasks, each image contains a single object on an unconstrained background.

\begin{itemize}[leftmargin=*]

  \item \textbf{CIFAR-10 \cite{krizhevsky2009learning}:} This dataset consists of $60,000$ (32 x 32) images belonging to 10 classes (airplane, automobile, bird, cat, deer, dog, frog, horse, ship, and truck), and is divided into $50,000$ images to be used for training and $10,000$ for testing. 
  
  \item \textbf{CIFAR-100 \cite{krizhevsky2009learning}:} Similar to CIFAR-10 dataset, CIFAR-100 consists of $50,000$ training images and $10,000$ testing images. There are $100$ classes in this dataset, grouped into 20 super-classes. Each image comes with a ``fine label" which is the class label and a ``coarse label" which is the super-class to which it belongs. In our study, we make use of the fine labels.

  \item \textbf{SVHN \cite{netzer2011reading}:} This real-world dataset comprises house numbers obtained from Google Street View images and consists of $73,257$ samples for training, $26,032$ images for testing and $531,131$ additional, less difficult samples which can be used as extra data for training. In our study, the training
  portion of the dataset corresponding to isolated digits (10 classes) is used.
  
   \item \textbf{PlantVillage \cite{hughes2015open} :} This crowd-sourced dataset consists of $54,309$ images with $39$ diseases of different crop plants. Three different versions of this dataset are provided: original \mbox{RGB} images with varied sizes, gray-scaled version of the raw images, and \mbox{RGB} images with just the leaf segmented and color corrected. In this work, we used the original \mbox{RGB} images. 
\end{itemize}

\subsection{Base Network}
\label{basenetworks}

To construct the base network, we employ three commonly used, state-of-the-art convolutional neural network architectures; namely Inception-V3 \cite{szegedy2016rethinking}, Xception \cite{chollet2017xception}, and Squeeze-and-Excitation Networks (SENet) \cite{hu2018squeeze}. 

Inception-V3, proposed by Google in \cite{szegedy2016rethinking}, is a widely-used image recognition model. It consists of symmetric and asymmetric building blocks, including convolutions, average pooling, max pooling, dropouts, and fully connected layers. Batch normalisation is used extensively throughout the model and applied to activation inputs. 
The network is $48$ layers deep with a total of $23.8$ million parameters.

Xception \cite{chollet2017xception} is an extreme version and an extension of the Inception \cite{szegedy2015going} architecture, which replaces the standard inception modules with depth-wise separable convolutions. The network is $71$ layers deep with a total of $22.9$ million parameters.

The final architecture, SENet \cite{hu2018squeeze}, introduces the squeeze-and-excitation block that adaptively re-calibrates the channel-wise feature responses by modelling the interdependencies between channels to automatically acquire the importance of each feature channel. SENet is the winner of ILSVRC 2017 classification challenge.


\subsection{Experimental Setting}\label{expset}

All the base networks employed in this study are pre-trained on the ImageNet dataset \cite{deng2009imagenet}. The inputs to the ensemble systems are obtained by forward passing the datasets through the fine-tuned networks and extracting the features of the last pooling layer ($2048$ neurons). 
Note that for a fair comparison, no extra randomness such as data augmentation, has been applied during training or feature extraction. 

The averaging ensembles are obtained by training 5 base networks with different random weight initialisations, as this is a typical number employed in these ensembles to respect time and computing power constraints.   
For randECOC using independently trained base classifiers, we train $L$ classifiers with a simple multi-layer perceptron architecture, where $L$ is set to $30$, $300$, and $30$ for CIFAR-10, CIFAR-100, and SVHN datasets, respectively. The architecture, which is depicted in Figure \ref{figindep}, consists of three fully connected layers with ($500$, $50$, $10$) units, each followed by rectified linear unit $(ReLU)$ activation function and a dropout layer. Finally, each output layer has one neuron that is associated with a tangent hyperbolic activation function $(tanh)$ and a mean square error $(MSE)$ loss function. 

The randECOC framework using MTL is composed of two shared fully connected layers with ($500$, $50$) units, each of which is followed by a $ReLU$ activation function and a dropout layer. For each classifier, there are some specific layers: a dropout layer, a fully connected layer with $10$ units, a $ReLU$ activation function, a fully connected layer with one unit, and a $tanh$ function.  The output units are concatenated to form one layer with $L$ units defining the output layer. Similar to randECOC with independently trained base classifiers, $MSE$ loss function is used here. The network structure of this framework is as given in Figure \ref{MTLbasic}. 

The randECOC using MTL w/ embedding, as given in Figure \ref{MTL-e2e}, mimics the setup of the randECOC framework using MTL, with an additional layer to include ECOC codewords as weights, for which the learning rate is set to zero. Note that as the random weight initialisations impacts on training all randECOC frameworks, we report the mean and the standard deviation of the testing accuracy from 5 independent runs.


All the networks including the base networks are optimised using RMSPROP optimiser with $3\times10^{-4}$ learning rate, $0.99$ squared gradient decay factor, and a batch size of $64$ images per training iteration for the base networks and $512$ images for the randECOC experiments. The implementation is performed using the Deep Learning Toolbox and MatConvNet \cite{vedaldi2015matconvnet} within MATLAB, with a single NVIDIA GeForce GTX 1080 Ti 11GB graphics processing unit (GPU).


The GBDT frameworks are implemented using the official XGBoost, LightGBM and CatBoost Python packages on Google Colaboratory with the provided free Tesla K80 11GB GPU. In our experiments, we fine-tune the most vital hyper-parameter for these frameworks, which is the number of iterations that is relative to the number of created trees. The highest validation accuracy has been obtained using $60$ iterations for CIFAR10 and SVHN datasets and $300$ iterations for CIFAR100 in all of the employed gradient boosting methods. Rest of the hyper-parameters are set to the default values suggested by the corresponding authors.

\begin{table*}[h]
\centering
\resizebox{\linewidth}{!}{
\begin{tabular}{c|c|c||c|c|c||c|c|c|}
\cline{1-9}
\multicolumn{3}{|c||}{}  & 
\multicolumn{3}{|c||}{\textbf{Gradient Boosting Ensemble}} & \multicolumn{3}{|c|}{\textbf{randECOC Ensemble}} \\ \hline

\multicolumn{1}{|c|}{}                                                                & \multicolumn{1}{c|}{\begin{tabular}[c]{@{}c@{}}Base \\ Network\end{tabular}} 
& \begin{tabular}[c]{@{}c@{}}Ensemble of 5\\base networks\end{tabular} & \small{XGBoost} &\small{LightGBM} & \small{CatBoost} & \begin{tabular}[c]{@{}c@{}}\small{Independent}\\ \small{Classifiers}\end{tabular} & \begin{tabular}[c]{@{}c@{}}\small{~~MTL~~}\\\end{tabular} & \begin{tabular}[c]{@{}c@{}}\small{MTL w/} \\ \small{embedding}\end{tabular} \\ \hline

\multicolumn{9}{|c|}{\textbf{Inception-V3}} \\ \hline
\multicolumn{1}{|c|}{\begin{tabular}[c]{@{}c@{}}Testing\\ Accuracy\end{tabular}} & 93.56\% &  
\begin{tabular}[c]{@{}c@{}} 96.14\% \\ $(+ 2.58)$\end{tabular} 
& \begin{tabular}[c]{@{}c@{}} \textbf{94.39\%} \\ $(+ 0.83)$\end{tabular} 
& \begin{tabular}[c]{@{}c@{}} 94.03\% \\ $(+ 0.47)$\end{tabular} 
& \begin{tabular}[c]{@{}c@{}} {93.37\%} \\ $(- 0.19)$\end{tabular}   &  
\begin{tabular}[c]{@{}c@{}} \textbf{94.61\% $\pm0.027$} \\ $(+ 1.05)$\end{tabular}& 
\begin{tabular}[c]{@{}c@{}} 94.45\% {$\pm0.033$}\\ $(+ 0.89)$\end{tabular}& 
\begin{tabular}[c]{@{}c@{}} 94.49\% {$\pm0.087$}\\ $(+ 0.93)$\end{tabular}\\ \hline
\multicolumn{1}{|c|}{\begin{tabular}[c]{@{}c@{}}Training Time\\ (in minutes)\end{tabular}} & 580 & 2,900 & 0.78 & 4.25  & 0.36 & 103 & 4.27  &  4.63\\ \hline
\multicolumn{1}{|c|}{\begin{tabular}[c]{@{}c@{}}Testing Time\\ (in minutes)\end{tabular}}  & 1.70 & 8.50 & 0.002  & 0.006 & 0.021 &  0.130  &  0.022 &  0.022\\ \hline

\multicolumn{9}{|c|}{\textbf{Xception}} \\ \hline
\multicolumn{1}{|c|}{\begin{tabular}[c]{@{}c@{}}Testing\\ Accuracy\end{tabular}} & 94.88\%  
& \begin{tabular}[c]{@{}c@{}} 97.02\% \\ $(+ 2.14)$\end{tabular}
& \begin{tabular}[c]{@{}c@{}} \textbf{95.18\%} \\ $(+ 0.30)$\end{tabular}
& \begin{tabular}[c]{@{}c@{}} 95.13\% \\ $(+ 0.25)$\end{tabular}
& \begin{tabular}[c]{@{}c@{}} 94.98\% \\ $(+ 0.10)$\end{tabular} 
& 
\begin{tabular}[c]{@{}c@{}} \textbf{95.52\%$\pm0.044$}\\ $(+ 0.64)$\end{tabular}& 
\begin{tabular}[c]{@{}c@{}} 95.40\%$\pm0.062$\\ $(+ 0.52)$\end{tabular}& 
\begin{tabular}[c]{@{}c@{}} 95.42\%$\pm0.048$\\ $(+ 0.54)$\end{tabular}\\ \hline
\multicolumn{1}{|c|}{\begin{tabular}[c]{@{}c@{}}Training Time\\ (in minutes)\end{tabular}} & 722  & 3,611   & 0.76  &  4.31      & 0.36   &   103     &   4.27 &   4.63 \\ \hline
\multicolumn{1}{|c|}{\begin{tabular}[c]{@{}c@{}}Testing Time\\ (in minutes)\end{tabular}}  & 2.79  &   13.9    & 0.002 &  0.007     & 0.027     &  0.124    &   0.022 &    0.022  \\ \hline

\multicolumn{9}{|c|}{\textbf{SENet}} \\ \hline
\multicolumn{1}{|c|}{\begin{tabular}[c]{@{}c@{}}Test\\ Accuracy\end{tabular}} & 95.93\% & \begin{tabular}[c]{@{}c@{}} 97.69 \% \\ $(+ 1.76)$\end{tabular}  
& \begin{tabular}[c]{@{}c@{}} \textbf{96.33\%} \\ $(+ 0.40)$\end{tabular}   
& \begin{tabular}[c]{@{}c@{}} 96.12\% \\ $(+ 0.19)$\end{tabular} 
& \begin{tabular}[c]{@{}c@{}} {95.07\%} \\ $(- 0.86)$\end{tabular}  
& 
\begin{tabular}[c]{@{}c@{}} \textbf{96.82\% $\pm0.024$} \\ $(+ 0.89)$\end{tabular}& 
\begin{tabular}[c]{@{}c@{}} 96.81\%$\pm0.067$ \\ $(+ 0.88)$ \end{tabular}& 
\begin{tabular}[c]{@{}c@{}} 96.81\%$\pm0.074$ \\ $(+ 0.88)$ \end{tabular}\\ \hline
\multicolumn{1}{|c|}{\begin{tabular}[c]{@{}c@{}}Training Time\\ (in minutes)\end{tabular}} & 1430 &   7,154   &   0.780 & 4.58  &  0.36  &   103   &  4.27  &  4.62 \\ \hline
\multicolumn{1}{|c|}{\begin{tabular}[c]{@{}c@{}}Testing Time\\ (in minutes)\end{tabular}}  & 6.98  &   34.92   &   0.002 & 0.007 &  0.026  &   0.129 &  0.022 &  0.022  \\ \hline

\end{tabular}
}
\caption{Comparison of the results obtained on the CIFAR-10 dataset using Inceptions-V3, Xception, and SENet architectures as base networks. The best results obtained in each group are shown in bold and the performance decreases compared to the base networks are shown underlined. The numbers in parentheses show the performance change compared to the base network}.
\label{tbl:CIFAR10_EXP}
\end{table*}


For the set of experiments where the randECOC and GBDT frameworks are combined with ensemble averaging, we train 5 base networks, each of which is initialised by random weights, separately for all  network architectures and dataset combinations. For each network architecture, we first evaluate the ensemble averaging performance with the 5 networks. Then, GBDT and randECOC approaches are applied to the features extracted from each base network, resulting in 5 ensembles in each case. The ECOC matrices used in all 5 networks are kept the same. Finally, ensemble averaging is applied to the 5 GDBT and 5 randECOC ensembles to obtain the final prediction.


To further validate our results, we carry out a final set of experiments with the PlantVillage dataset which is a large, crowd-sourced dataset of real-life diseased plant images.

\subsection{Results}\label{results}

We report the result of a comparison of the evaluated frameworks in Section \ref{acTimeComp} and the performance analysis of combinatory approaches in Section \ref{ave_ecoc_gbdt}.

\subsubsection{Comparison of the Ensemble Frameworks} 
\label{acTimeComp}
The performance of the 3 ensemble frameworks together with their corresponding time complexity, while using three base networks, is shown in Table \ref{tbl:CIFAR10_EXP}, \ref{tbl:CIFAR100_EXP}, and \ref{tbl:SVHN_EXP} for CIFAR-10, CIFAR-100 and SVHN datasets, respectively. The performance is gauged in terms of classification accuracy, and the time complexity is measured as the training and test time spent, \textit{over and above} the time required by the base network. Note that while the hardware is slightly different for GBDT and ECOC frameworks, their time complexities are both accepted as small and no strict comparison is made between the two in terms of time.

\textbf{Ensemble Averaging:} As expected, the averaging ensemble achieves the highest accuracy for all datasets and base networks, 
with highest accuracies of
97.69\%, 89.91\%  and 97.75\%  for the CIFAR-10, CIFAR-100 and SVHN datasets, respectively. Despite surpassing the base network by a relatively high margin, this ensemble comes out as very costly in terms of time and the resources required. Specifically, the training times are of the order of thousands of minutes, or about several days, which is often not available to researchers, providing the motivation for this work. 

\textbf{Gradient Boosting Decision Trees:} Among the GBDT frameworks, 
none outperforms the others on all datasets, and more importantly, it can be observed that all three variations cause degradation over the base network performance at least for network architecture and dataset. This is a very important finding, proving a clear evidence in support of the perceived instability and inconsistency of this technique, especially when dealing with a high number of classes. 
Specifically, for CIFAR-10 and SVHN datasets, XGBoost appears as the best performing algorithm as shown in Table \ref{tbl:CIFAR10_EXP} and \ref{tbl:SVHN_EXP}, respectively. It improves the testing accuracy of the base networks at the expense of minimal additional training time, with improvements of $0.83\%$, $0.30\%$, and $0.40\%$ on CIFAR-10 and $0.89\%$, $0.20\%$, and $0.19\%$ on the SVHN dataset, compared to the base network. 
However, this method deteriorates the base network accuracy for all network types on CIFAR-100. For this dataset, the only GBDT improvement over the base network performance is achieved when using Xception as the architecture and employing LightGBM or CatBoost. 
This is in line with the theoretical underpinning of the inability of these methods to cope with a high number of classes \cite{kuralenok2019factorized}. 
%

\begin{table*}[h]
\centering
\resizebox{\linewidth}{!}{
\begin{tabular}{c|c|c||c|c|c||c|c|c|}
\cline{1-9}
\multicolumn{3}{|c||}{}  &
\multicolumn{3}{|c||}{\textbf{Gradient Boosting Ensemble}} & \multicolumn{3}{|c|}{\textbf{randECOC Ensemble}} \\ \hline

\multicolumn{1}{|c|}{} & \multicolumn{1}{|c|}{\begin{tabular}[c]{@{}c@{}}Base\\ Network\end{tabular}} & \begin{tabular}[c]{@{}c@{}}Averaging of\\5 base networks\end{tabular}  & \small{XGBoost} &\small{LightGBM} & \small{CatBoost} & \begin{tabular}[c]{@{}c@{}}\small{Independent}\\ \small{Classifiers}\end{tabular} & \begin{tabular}[c]{@{}c@{}}\small{~~MTL~~}\\\end{tabular} & \begin{tabular}[c]{@{}c@{}}\small{MTL w/} \\ \small{embedding}\end{tabular} \\ \hline

\multicolumn{9}{|c|}{\textbf{Inception-V3}} \\ \hline
\multicolumn{1}{|c|}{\begin{tabular}[c]{@{}c@{}}Test\\ Accuracy\end{tabular}}  & 76.77\%   
& \begin{tabular}[c]{@{}c@{}} 81.34\% \\ $(+ 4.57)$\end{tabular}
& \begin{tabular}[c]{@{}c@{}} {73.67\%} \\ $(- 3.10)$\end{tabular}
& \begin{tabular}[c]{@{}c@{}} {\textbf{76.47\%}} \\ $(- 0.30)$\end{tabular} 
& \begin{tabular}[c]{@{}c@{}} {76.12\%} \\ $(- 0.65)$\end{tabular} & 

\begin{tabular}[c]{@{}c@{}}\textbf{78.91\%} $\pm0.025$ \\ $(+ 2.14 )$\end{tabular}& 
\begin{tabular}[c]{@{}c@{}}78.34\% $\pm0.091$ \\ $(+ 1.57)$\end{tabular}& 
\begin{tabular}[c]{@{}c@{}}78.45\% $\pm0.096$ \\ $(+ 1.68)$\end{tabular}\\ \hline

\multicolumn{1}{|c|}{\begin{tabular}[c]{@{}c@{}}Training Time\\ (in minutes)\end{tabular}} & 1160 
    &  5800    &  47.3  &   175    &   5.27 &  981 & 34.8 & 35.7   \\ \hline

\multicolumn{1}{|c|}{\begin{tabular}[c]{@{}c@{}}Testing Time\\ (in minutes)\end{tabular}}  & 1.66  &  8.28    &  0.012 &  0.344   &  0.024 &  1.25 & 0.187  & 0.191     \\\hline
  
\multicolumn{9}{|c|}{\textbf{Xception}} \\ \hline
\multicolumn{1}{|c|}{\begin{tabular}[c]{@{}c@{}}Test\\ Accuracy\end{tabular}}  & 80.67\%  
& \begin{tabular}[c]{@{}c@{}} 85.50\% \\ $+ 4.83)$\end{tabular}
& \begin{tabular}[c]{@{}c@{}} {79.30\%} \\ $(- 1.37)$\end{tabular}
& \begin{tabular}[c]{@{}c@{}} \textbf{81.70\%} \\ $(+ 1.03)$\end{tabular} 
& \begin{tabular}[c]{@{}c@{}} 81.58\% \\ $(+ 0.91)$\end{tabular}& 
\begin{tabular}[c]{@{}c@{}}\textbf{83.24\%} $\pm0.035$ \\ $(+ 2.57)$\end{tabular}& 
\begin{tabular}[c]{@{}c@{}}82.97\% $\pm0.077$ \\ $(+ 2.30)$\end{tabular}& 
\begin{tabular}[c]{@{}c@{}}83.16\% $\pm0.081$ \\ $(+ 2.49)$\end{tabular}\\ \hline
\multicolumn{1}{|c|}{\begin{tabular}[c]{@{}c@{}}Training Time\\ (in minutes)\end{tabular}} 
& 1342 &  6709   & 47.7  &  178  &  5.35  & 1025 & 38.2  &   38.4   \\ \hline
\multicolumn{1}{|c|}{\begin{tabular}[c]{@{}c@{}}Testing Time\\ (in minutes)\end{tabular}}  & 3.03 &  15.2 & 0.012 & 0.327 &  0.025 & 1.35 & 0.197 & 0.205\\ \hline

\multicolumn{9}{|c|}{\textbf{SENet}} \\ \hline
\multicolumn{1}{|c|}{\begin{tabular}[c]{@{}c@{}}Test\\ Accuracy\end{tabular}}   & 87.35\% 
& \begin{tabular}[c]{@{}c@{}} 89.91\% \\ $(+ 2.56)$\end{tabular} 
& \begin{tabular}[c]{@{}c@{}} {84.17\%} \\ $(- 3.18)$\end{tabular}
& \begin{tabular}[c]{@{}c@{}} {86.39\%} \\ $(- 0.96)$\end{tabular}
& \begin{tabular}[c]{@{}c@{}} {\textbf{86.51\%}} \\ $(- 0.84)$\end{tabular} & 
\begin{tabular}[c]{@{}c@{}} \textbf{87.90\%} $\pm0.040$ \\ $(+ 0.55)$\end{tabular}& 
\begin{tabular}[c]{@{}c@{}} 87.60\% $\pm0.037$ \\ $(+ 0.25)$\end{tabular}& 
\begin{tabular}[c]{@{}c@{}} 87.74\% $\pm0.130$ \\ $(+ 0.39)$\end{tabular}\\ \hline
\multicolumn{1}{|c|}{\begin{tabular}[c]{@{}c@{}}Training Time\\ (in minutes)\end{tabular}} 
& 1463 &  7317  &  48.4 & 179  &  5.57  & 1005   &  35.8 &  36.4 \\ \hline
\multicolumn{1}{|c|}{\begin{tabular}[c]{@{}c@{}}Testing Time\\ (in minutes)\end{tabular}} & 7.67 &   38   & 0.007 & 0.391  &  0.045 & 1.28   & 0.195 &  0.213\\ \hline
\end{tabular}
}
\caption{Comparisons on the CIFAR-100 dataset using the Inceptions-V3, Xception, and SENet architectures as base networks. The best results obtained in each group are shown in bold.
}
\label{tbl:CIFAR100_EXP}
\end{table*}

\begin{table*}[h]
\centering
\resizebox{\linewidth}{!}{
\begin{tabular}{c|c|c||c|c|c||c|c|c|}
\cline{1-9}
\multicolumn{3}{|c||}{}  &
\multicolumn{3}{|c||}{\textbf{Gradient Boosting Ensemble}} & 
\multicolumn{3}{|c|}{\textbf{randECOC Ensemble}} \\ \hline
\multicolumn{1}{|c|}{}                                                                     & \multicolumn{1}{c|}{\begin{tabular}[c]{@{}c@{}}Base\\ Network\end{tabular}} & \begin{tabular}[c]{@{}c@{}}Averaging of\\5 base networks\end{tabular} & \small{XGBoost} &\small{LightGBM} & \small{CatBoost} & \begin{tabular}[c]{@{}c@{}}\small{Independent}\\ \small{Classifiers}\end{tabular} & \begin{tabular}[c]{@{}c@{}}\small{~~MTL~~}\\\end{tabular} & \begin{tabular}[c]{@{}c@{}}\small{MTL w/} \\ \small{embedding}\end{tabular} \\ \hline

\multicolumn{9}{|c|}{\textbf{Inception-V3}} \\ \hline
\multicolumn{1}{|c|}{\begin{tabular}[c]{@{}c@{}}Test\\ Accuracy\end{tabular}}          
& 95.63\% 
& \begin{tabular}[c]{@{}c@{}} 97.51\% \\ $(+ 1.88)$\end{tabular} 
& \begin{tabular}[c]{@{}c@{}} \textbf{96.52\%} \\ $(+ 0.89)$\end{tabular}  
& \begin{tabular}[c]{@{}c@{}} 96.42\% \\ $(+ 0.79)$\end{tabular} 
& \begin{tabular}[c]{@{}c@{}} 96.20\% \\ $(+ 0.57)$\end{tabular} 
&  
\begin{tabular}[c]{@{}c@{}} \textbf{96.76\%} $\pm0.018$ \\ $(+ 1.13)$\end{tabular}& 
\begin{tabular}[c]{@{}c@{}} 96.67\% $\pm0.017$ \\ $(+ 1.04)$\end{tabular}& 
\begin{tabular}[c]{@{}c@{}} 96.67\% $\pm0.045$ \\ $(+ 1.04)$\end{tabular}\\ \hline
\multicolumn{1}{|c|}{\begin{tabular}[c]{@{}c@{}}Training Time\\ (in minutes)\end{tabular}} 
& 690 &   3,450   & 0.820  & 1.57 &  0.570   & 151      & 6.49   & 6.81  \\ \hline
\multicolumn{1}{|c|}{\begin{tabular}[c]{@{}c@{}}Testing Time\\ (in minutes)\end{tabular}}  
& 4.42 &  22.1  & 0.004  & 0.008 & 0.035   & 0.335    & 0.064  & 0.065  \\ \hline

\multicolumn{9}{|c|}{\textbf{Xception}} \\ \hline
\multicolumn{1}{|c|}{\begin{tabular}[c]{@{}c@{}}Test\\ Accuracy\end{tabular}}           & 96.83\%  
& \begin{tabular}[c]{@{}c@{}} 97.75\% \\ $(+ 0.92)$\end{tabular}
& \begin{tabular}[c]{@{}c@{}} \textbf{97.03\%} \\ $(+ 0.20)$\end{tabular}
& \begin{tabular}[c]{@{}c@{}} 97.02\% \\ $(+ 0.19)$\end{tabular}
& \begin{tabular}[c]{@{}c@{}} 96.98\% \\ $(+ 0.15)$\end{tabular}
& 
\begin{tabular}[c]{@{}c@{}} \textbf{97.15\%} $\pm0.011$ \\ $(+ 0.32)$\end{tabular}& 
\begin{tabular}[c]{@{}c@{}} 97.10\% $\pm0.010$ \\ $(+ 0.27)$ \end{tabular}& 
\begin{tabular}[c]{@{}c@{}} 97.12\% $\pm0.010$ \\ $(+ 0.29)$ \end{tabular}\\ \hline
\multicolumn{1}{|c|}{\begin{tabular}[c]{@{}c@{}}Training Time\\ (in minutes)\end{tabular}} 
& 830  &  4,150 &   1.14   & 2.28  &  0.34  &  153 & 6.53   &  6.85        \\ \hline
\multicolumn{1}{|c|}{\begin{tabular}[c]{@{}c@{}}Testing Time\\ (in minutes)\end{tabular}}  
& 7.59 &  37.9  &  0.004 & 0.007 & 0.037 &  0.342 & 0.063   &  0.064      \\ \hline

\multicolumn{9}{|c|}{\textbf{SENet}} \\ \hline
\multicolumn{1}{|c|}{\begin{tabular}[c]{@{}c@{}}Test\\ Accuracy\end{tabular}}           & 94.70\% 
& \begin{tabular}[c]{@{}c@{}} 95.81\% \\ $(+ 1.11)$\end{tabular}
& \begin{tabular}[c]{@{}c@{}} \textbf{94.89\% }\\ $(+ 0.19)$\end{tabular}
& \begin{tabular}[c]{@{}c@{}} 94.36\% \\ $(+ 0.34)$\end{tabular}
& \begin{tabular}[c]{@{}c@{}} {92.85\%} \\ $(- 1.85)$\end{tabular}
& 
\begin{tabular}[c]{@{}c@{}} \textbf{95.81\%} $\pm0.099$ \\ $(+ 1.11)$\end{tabular}& 
\begin{tabular}[c]{@{}c@{}} 95.72\% $\pm0.015$ \\ $(+ 1.02)$\end{tabular}& 
\begin{tabular}[c]{@{}c@{}} 95.79\% $\pm0.038$ \\ $(+ 1.09)$\end{tabular}\\ \hline
\multicolumn{1}{|c|}{\begin{tabular}[c]{@{}c@{}}Training Time\\ (in minutes)\end{tabular}} & 
1,877 &  9,594  &  1.14  & 2.34   & 0.49  &   151     &  6.58 &   6.79  \\ \hline
\multicolumn{1}{|c|}{\begin{tabular}[c]{@{}c@{}}Testing Time\\ (in minutes)\end{tabular}}  & 
18.2  &  90.5  &  0.005 & 0.007  & 0.043 &   0.402   &  0.061 &  0.065  \\ \hline

  \end{tabular}
}
\caption{Comparisons on the SVHN dataset using the Inceptions-V3, Xception, and SENet architectures as base networks. The best results obtained in each group are shown in bold.
}
\label{tbl:SVHN_EXP}
\end{table*}


\textbf{randECOC Ensembles:} We see that all the variants of the randECOC framework improve the testing accuracy over the base networks and the best GBDT approach\footnote{except for one out of the nine settings, where a slight drop for the MTL approach was noted.}, in almost all of our experiments. 
Despite some drop in the performance in comparison to the averaging ensembles, a much faster training time is observed. For instance, in the case of CIFAR-100, the averaging ensemble requires around $4$, $4.6$, $5$ days for training and reveals $81.34\%$, $85.50\%$, $89.91\%$ test accuracy, with different base network architectures. On the other hand, randECOC using MTL w/ embedding requires only about $30$ minutes for the training of all the architectures, with the output test accuracy of $78.45\%$, $83.16\%$, and $87.74\%$.
While MTL w/ embedding brings roughly half the performance improvement obtained by the averaging ensemble over the base network, 
it does so consistently and requiring negligible additional time, which is important  for scenarios where training several deep networks is not viable.

Among the MTL based randECOC ensembles, MTL w/ embedding performs always better than or equal to MTL, while revealing similar time complexity. The independent learning approach obtains the highest accuracy; however only with a slight margin over MTL w/ embedding and a lot more additional training time  (more than 20 times in all scenarios). 


\begin{table*}[h]
\centering
\resizebox{\linewidth}{!}{
\begin{tabular}{c|c|c||c|c|c||c|c|c|}
\cline{1-9}
\multicolumn{3}{|c||}{}  & 
\multicolumn{3}{|c||}{\textbf{Gradient Boosting Ensemble}} & \multicolumn{3}{|c|}{\textbf{randECOC Ensemble}} \\ \hline
\multicolumn{1}{|c|}{}                                                                     & \multicolumn{1}{c|}{\begin{tabular}[c]{@{}c@{}}Base\\ Network\end{tabular}} & \begin{tabular}[c]{@{}c@{}}Averaging of\\5 base networks\end{tabular} & \small{XGBoost} &\small{LightGBM} & \small{CatBoost} & \begin{tabular}[c]{@{}c@{}}\small{Independent}\\ \small{Classifiers}\end{tabular} & \begin{tabular}[c]{@{}c@{}}\small{~~MTL~~}\\\end{tabular} & \begin{tabular}[c]{@{}c@{}}\small{MTL w/} \\ \small{embedding}\end{tabular} \\ \hline

\multicolumn{9}{|c|}{\textbf{CIFAR-10}} \\ \hline
\multicolumn{1}{|c|}{Inception-V3}
&  93.56\%  & 96.14\% & 96.22\% &  96.25\% &  96.03\% &  \textbf{96.42\%} & 96.35\% & 96.35\% \\ \hline
\multicolumn{1}{|c|}{Xception}    
&  94.88\%  & 97.02\% & 97.09\% &  97.06\% &  97.11\% &  \textbf{97.25\%} & 97.18\% & 97.20\% \\ \hline
\multicolumn{1}{|c|}{SENet}       
&  95.93\%  & 97.69\% & 97.70\% &  97.77\% &  97.16\% &  \textbf{97.85\%} & 97.79\% & 97.84\% \\ \hline

\multicolumn{9}{|c|}{\textbf{CIFAR-100}} \\ \hline
\multicolumn{1}{|c|}{Inception-V3}
&  76.77\%  & 81.34\% & 81.50\% &  82.50\% &  81.68\% &  \textbf{83.34\%} & 83.12\% & 83.26\% \\ \hline
\multicolumn{1}{|c|}{Xception}    
&  80.67\%  & 85.50\% & 85.59\% &  85.88\% &  85.95\% &  \textbf{86.65\%} & 86.51\% & 86.55\% \\ \hline
\multicolumn{1}{|c|}{SENet}       
&  87.35\%  & 89.91\% & 88.87\% &  89.30\% &  88.44\% &  \textbf{90.11\%} & 90.00\% & 90.07\% \\ \hline

\multicolumn{9}{|c|}{\textbf{SVHN}} \\ \hline
\multicolumn{1}{|c|}{Inception-V3}
&  95.63\%  &  97.51\% & 97.57\% &  97.41\% &  97.56\% &  \textbf{97.59\%} & 97.55\% & 97.56\% \\ \hline
\multicolumn{1}{|c|}{Xception}   
&  96.83\%  &  97.75\% & 97.74\% &  97.71\% &  97.72\% &  \textbf{97.80\%} & 97.75\% & 97.78\% \\ \hline
\multicolumn{1}{|c|}{SENet}       
&  94.70\%  &  95.81\% & 96.14\% &  96.02\% &  95.67\% &  \textbf{96.27\%} & 96.16\% & 96.22\% \\ \hline

\end{tabular}
}
\caption{Test accuracies for the combinatory methods. 
The best result corresponding to each dataset and base network, is shown in bold.
}
\label{tbl:EnsembleExperiments}
\end{table*}

\begin{table*}[h]
\centering
\resizebox{\linewidth}{!}
{
\begin{tabular}{|c|c|c|c|c||c|c|c|}
\cline{1-8}
\multicolumn{2}{|c|}{}  &
\multicolumn{3}{|c|}{\textbf{Gradient Boosting Ensemble}} & 
\multicolumn{3}{|c|}{\textbf{randECOC Ensemble}} \\ \hline

\multicolumn{1}{|c|}{} & \multicolumn{1}{|c|}{\begin{tabular}[c]{@{}c@{}}Base\\ Network\end{tabular}} &\small{XGBoost} &\small{LightGBM} & \small{CatBoost} & \begin{tabular}[c]{@{}c@{}}\small{Independent}\\ \small{Classifiers}\end{tabular} & \begin{tabular}[c]{@{}c@{}}\small{~~MTL~~}\\\end{tabular} & \begin{tabular}[c]{@{}c@{}}\small{MTL w/} \\ \small{embedding}\end{tabular} \\ \hline

\multicolumn{1}{|c|}{~~~~Fold-1~~~~}&  
\multicolumn{1}{c|}{99.52\%} & 99.32\% &  \textbf{99.41\%} &  \textbf{99.41\%} &
\textbf{99.71\%} $\pm0.008$& 99.68\% $\pm0.011$& 99.68\% $\pm0.009$\\ \hline
\multicolumn{1}{|c|}{Fold-2}&  99.60\% & 99.53\%  &  99.46\% &  \textbf{99.54\%} & 
\textbf{99.73\%} $\pm0.014$&99.66\% $\pm0.009$&99.68\% $\pm0.011$\\ \hline
\multicolumn{1}{|c|}{Fold-3}&  99.44\% &  99.36\%  &  99.27\% &  \textbf{99.41\%} & 
\textbf{99.70\%} $\pm0.012$&99.59\% $\pm0.013$&99.61\% $\pm0.008$\\ \hline
\multicolumn{1}{|c|}{Fold-4} & 99.38\% &  99.36\%   &  \textbf{99.41\%} &  99.41\% & 
\textbf{99.67\%} $\pm0.011$&99.60\% $\pm0.012$&99.64\% $\pm0.014$\\ \hline
\multicolumn{1}{|c|}{Fold-5}& 99.60\% &  99.40\% &  99.32\% &  \textbf{99.52\%} & 
\textbf{99.71\%} $\pm0.009$&99.64\% $\pm0.011$&99.68\% $\pm0.008$\\ \hline
\multicolumn{1}{|c|}{Average} & 95.51\% & 99.40\% &  99.37\% &  \textbf{99.46\%} & 
\textbf{99.70\%} & 99.63\% & 99.66\% \\\hline
\cline{1-8}
\multicolumn{8}{|c|}{Combinatory Approach}\\ \hline
\multicolumn{1}{|c|}{} & \multicolumn{1}{|c|}{\begin{tabular}[c]{@{}c@{}}Averaging of\\ 5 base networks\end{tabular}} &\small{XGBoost} &\small{LightGBM} & \small{CatBoost} & \begin{tabular}[c]{@{}c@{}}\small{Independent}\\ \small{Classifiers}\end{tabular} & \begin{tabular}[c]{@{}c@{}}\small{~~MTL~~}\\\end{tabular} & \begin{tabular}[c]{@{}c@{}}\small{MTL w/} \\ \small{embedding}\end{tabular} \\ \hline
\multicolumn{1}{|c|}{Fold 1} & 99.76\% & 99.72\% &  99.74\% &  99.72\% &
\textbf{99.81\%} & 99.79\% & 99.79\%\\ \hline


\end{tabular}
}
\caption{5-Fold cross validation and the combinatory approach on the Plant Village dataset using the Xception base network. The best results obtained in each group are shown in bold.
}
\label{tbl:PlantVillage}
\end{table*}

The strength of the MTL based randECOC approaches over GBDTs is emphasised especially when dealing with high number of classes. As shown in Table \ref{tbl:CIFAR100_EXP} for the CIFAR-100, MTL w/ embedding improves the  accuracy by $1.65\%$, $2.49\%$, $0.55\%$ over the base networks, and outperforms the best GBDT approach
(LightGBM in this case) by $1.98\%$, $1.46\%$, $1.35\%$, for the three network architectures. Note also that, the training time of LightGBM for this problem is also greater than that of MTL w/ embedding.

\subsubsection{Combinatory Approach - Ensemble Averaging of GBDT and randECOC Ensembles }  \label{ave_ecoc_gbdt}

As an important outcome of the comparative experiments presented in Section \ref{acTimeComp}, the averaging ensembles tend to achieve the highest accuracy for all the base networks and dataset combinations, benefiting from their increased computational complexity. Under the assumption of an adequate computational resources, we aim further to improve this accuracy by assisting the averaging process with GBDT and randECOC, as explained in \ref{expset}.
%

The results of these  experiments are provided in Table \ref{tbl:EnsembleExperiments}.
It can be observed that GBDT+averaging approaches outperform the baseline averaging ensemble by the slightest margin, while the randECOC+averaging methods provide a higher performance improvement, ranging from $0.05$ up to $2$ percentage points, where the highest improvement is observed for the CIFAR-100 dataset.

Although the best accuracies are acquired from randECOC using independent classifiers, MTL based approaches follow closely, revealing better accuracy than GBDTs in all scenarios other than one (Inception-V3 with SVHN), where the difference in performance with the best GBDT framework (XGBoost) is as small as $0.02\%$. 
The consistency in the improvement in accuracy not only over the base network, but also the baseline averaging ensemble and the GBDT+averaging ensemble, renders  randECOC+averaging as the best performing classifier combination technique in the literature.

We would like to underline the fact that the GBDT and randECOC frameworks operate on the features extracted by the base networks; hence training the combinatory approach with these frameworks takes little additional time.  
For instance, training 5 randECOC ensembles on top of the 5 base networks only takes 21 minutes for the CIFAR-10 dataset, while training the 5 base networks takes 3160 minutes. The additional time corresponds to 0.58\% overhead.

\subsubsection{Experiments with Real-Life  PlantVillage Dataset } \label{plantvillage}

Experiments on PlantVillage dataset \cite{hughes2015open} are done using 5-fold cross-validation due to the lack of a designated test set. 
The experiments are conducted only for GBDT and randECOC approaches using the Xception network due to its favorable performance-size ratio and the combinatory approach is applied on only one fold, due to large computational requirements.

Results on this dataset are shown in Table \ref{tbl:PlantVillage}, where
it can be observed that while all performances are very close, all randECOC variants achieve superior accuracy than all GBDT variants in all folds. 
Moreover, the combinatory approach of randECOC achieves the state-of-the-art results (99.81\%) on this dataset, while Mohanty et al., Too et al., and KC et al. reported \%99.34, \%99.75, and \%98.34 respectively \cite{mohanty2016using,too2019comparative,kamal2019depthwise}.
\section{Conclusions}\label{conc}

In this paper, we have proposed different design methodologies to address the use of the Error Correcting Output Coding (ECOC) framework as a strategy for constructing deep convolutional neural network ensembles. This is the first study to date, which comprehensively analyses ECOC in relation to the deep learning research, while proposing novel strategies to focus on the accuracy-complexity trade-off.

The current state-of-the-art deep ensemble techniques in the literature are constructed either by averaging the outputs of the multiple realisations of a deep network architecture by randomising / changing some of its constitutional elements, or by employing gradient boosting decision trees (GBDT) on the features extracted from one fully trained network. Despite all its advantages in terms of the performance gain, the increased time complexity the averaging ensembles incur, which is shown to be in the order of days and weeks for problems involving a high number of classes, may make this method computationally infeasible or inefficient for users with limited resources. Even though GBDTs  address this inefficiency, they have been shown to be unstable in terms of the improvement they offer over the base networks. In our experiments, we have shown that there exists no GBDT method which provides consistent improvement over the base accuracy for all architectures and datasets.

Addressing the drawbacks of GBDTs, we have proposed and analysed three ECOC-based design techniques, which provide a reliable and stable improvement over the base network performance as well as the performance of GBDT under all settings. Moreover, two of the proposed designs achieve time complexity benefits similar to GBDTs.

The proposed design techniques are based on independent learning, multi-task learning (MTL)
and multi-task learning with embedding (MTL w/ embedding). It has been shown that MTL w/ embedding always provides an accuracy equal to or greater than that of MTL, and both methods have a comparable time complexity with those of GBDTs.
Independent learning provides the best performance among the ECOC based methods. However, the performance gain over the MTL based methods is marginal and comes with the a time complexity trade-off, though this complexity is still much less than that of averaging. Therefore, for problems to be tackled with a limited  computational resource, we suggest that employing ECOC methods, the choice of which is to be made by the user depending on the fine-tuned requirements of the problem, is the best strategy; i.e. MTL w/ embedding for fastest training, independent learning for a relatively slower but marginally better performance. 

To offer solutions for scenarios where the available resources are not a limiting factor for the user, we have conducted experiments with simple averaging ensembles of GBDT and ECOC frameworks, and shown that the combinatory framework built using any of the ECOC methodologies achieves the best performances among all methods, at the expense of negligible additional training time. 

In conclusion, the ECOC framework, either alone or in combination with the averaging methodology, appears to provide the most efficient ensemble learning approach.In the future, the feasibility of end-to-end training of the proposed design strategies using the ECOC framework will be explored for the cases where time and space complexity is not a restriction.

\bibliographystyle{IEEEtran}  
\bibliography{main}

\begin{thebibliography}{10}
\providecommand{\url}[1]{#1}
\csname url@samestyle\endcsname
\providecommand{\newblock}{\relax}
\providecommand{\bibinfo}[2]{#2}
\providecommand{\BIBentrySTDinterwordspacing}{\spaceskip=0pt\relax}
\providecommand{\BIBentryALTinterwordstretchfactor}{4}
\providecommand{\BIBentryALTinterwordspacing}{\spaceskip=\fontdimen2\font plus
\BIBentryALTinterwordstretchfactor\fontdimen3\font minus
  \fontdimen4\font\relax}
\providecommand{\BIBforeignlanguage}[2]{{%
\expandafter\ifx\csname l@#1\endcsname\relax
\typeout{** WARNING: IEEEtran.bst: No hyphenation pattern has been}%
\typeout{** loaded for the language `#1'. Using the pattern for}%
\typeout{** the default language instead.}%
\else
\language=\csname l@#1\endcsname
\fi
#2}}
\providecommand{\BIBdecl}{\relax}
\BIBdecl

\bibitem{dietterich2000ensemble}
T.~G. Dietterich, ``Ensemble methods in machine learning,'' in
  \emph{International workshop on multiple classifier systems}.\hskip 1em plus
  0.5em minus 0.4em\relax Springer, 2000, pp. 1--15.

\bibitem{kuncheva2003measures}
L.~I. Kuncheva and C.~J. Whitaker, ``Measures of diversity in classifier
  ensembles and their relationship with the ensemble accuracy,'' \emph{Machine
  learning}, vol.~51, no.~2, pp. 181--207, 2003.

\bibitem{zhou2012ensemble}
Z.-H. Zhou, \emph{Ensemble methods: foundations and algorithms}.\hskip 1em plus
  0.5em minus 0.4em\relax CRC press, 2012.

\bibitem{polikar2012ensemble}
R.~Polikar, ``Ensemble learning,'' in \emph{Ensemble machine learning}.\hskip
  1em plus 0.5em minus 0.4em\relax Springer, 2012, pp. 1--34.

\bibitem{penrose1946elementary}
L.~S. Penrose, ``The elementary statistics of majority voting,'' \emph{Journal
  of the Royal Statistical Society}, vol. 109, no.~1, pp. 53--57, 1946.

\bibitem{breiman1996bagging}
L.~Breiman, ``Bagging predictors,'' \emph{Machine learning}, vol.~24, no.~2,
  pp. 123--140, 1996.

\bibitem{dvzeroski2004combining}
S.~D{\v{z}}eroski and B.~{\v{Z}}enko, ``Is combining classifiers with stacking
  better than selecting the best one?'' \emph{Machine learning}, vol.~54,
  no.~3, pp. 255--273, 2004.

\bibitem{breiman2001random}
L.~Breiman, ``Random forests,'' \emph{Machine learning}, vol.~45, no.~1, pp.
  5--32, 2001.

\bibitem{kong1995error}
E.~B. Kong and T.~G. Dietterich, ``Error-correcting output coding corrects bias
  and variance,'' in \emph{Machine Learning Proceedings 1995}.\hskip 1em plus
  0.5em minus 0.4em\relax Elsevier, 1995, pp. 313--321.

\bibitem{guruswami1999multiclass}
V.~Guruswami and A.~Sahai, ``Multiclass learning, boosting, and
  error-correcting codes,'' in \emph{Proceedings of the twelfth annual
  conference on Computational learning theory}, 1999, pp. 145--155.

\bibitem{deepface2014}
Y.~Taigman, M.~Yang, M.~Ranzato, and L.~Wolf, ``Deepface: Closing the gap to
  human-level performance in face verification,'' in \emph{Conference on
  Computer Vision and Pattern Recognition (CVPR)}, 2014.

\bibitem{szegedy2015going}
C.~Szegedy, W.~Liu, Y.~Jia, P.~Sermanet, S.~Reed, D.~Anguelov, D.~Erhan,
  V.~Vanhoucke, and A.~Rabinovich, ``Going deeper with convolutions,'' in
  \emph{Proceedings of the IEEE conference on computer vision and pattern
  recognition}, 2015, pp. 1--9.

\bibitem{redmon2016you}
J.~Redmon, S.~Divvala, R.~Girshick, and A.~Farhadi, ``You only look once:
  Unified, real-time object detection,'' in \emph{Proceedings of the IEEE
  conference on computer vision and pattern recognition}, 2016, pp. 779--788.

\bibitem{xiao2018deep}
Y.~Xiao, J.~Wu, Z.~Lin, and X.~Zhao, ``A deep learning-based multi-model
  ensemble method for cancer prediction,'' \emph{Computer methods and programs
  in biomedicine}, vol. 153, pp. 1--9, 2018.

\bibitem{gessert2020skin}
N.~Gessert, M.~Nielsen, M.~Shaikh, R.~Werner, and A.~Schlaefer, ``Skin lesion
  classification using ensembles of multi-resolution efficientnets with meta
  data,'' \emph{MethodsX}, p. 100864, 2020.

\bibitem{badjatiya2017deep}
P.~Badjatiya, S.~Gupta, M.~Gupta, and V.~Varma, ``Deep learning for hate speech
  detection in tweets,'' in \emph{Proceedings of the 26th International
  Conference on World Wide Web Companion}, 2017, pp. 759--760.

\bibitem{ren2017novel}
X.~Ren, H.~Guo, S.~Li, S.~Wang, and J.~Li, ``A novel image classification
  method with \mbox{CNN-XGBoost} model,'' in \emph{International Workshop on
  Digital Watermarking}.\hskip 1em plus 0.5em minus 0.4em\relax Springer, 2017,
  pp. 378--390.

\bibitem{pang2019novel}
L.~Pang, J.~Wang, L.~Zhao, C.~Wang, and H.~Zhan, ``A novel protein subcellular
  localization method with \mbox{CNN-XGBoost} model for alzheimer's disease,''
  \emph{Frontiers in genetics}, vol.~9, p. 751, 2019.

\bibitem{ju2019model}
Y.~Ju, G.~Sun, Q.~Chen, M.~Zhang, H.~Zhu, and M.~U. Rehman, ``A model combining
  convolutional neural network and \mbox{LightGBM} algorithm for
  ultra-short-term wind power forecasting,'' \emph{IEEE Access}, vol.~7, pp.
  28\,309--28\,318, 2019.

\bibitem{lasseck2017image}
M.~Lasseck, ``Image-based plant species identification with deep convolutional
  neural networks.'' in \emph{CLEF (Working Notes)}, 2017.

\bibitem{tan2019efficientnet}
M.~Tan and Q.~V. Le, ``Efficientnet: Rethinking model scaling for convolutional
  neural networks,'' \emph{arXiv preprint arXiv:1905.11946}, 2019.

\bibitem{chen2016xgboost}
T.~Chen and C.~Guestrin, ``\mbox{XGBoost}: A scalable tree boosting system,''
  in \emph{Proceedings of the 22nd acm sigkdd international conference on
  knowledge discovery and data mining}, 2016, pp. 785--794.

\bibitem{ke2017lightgbm}
G.~Ke, Q.~Meng, T.~Finley, T.~Wang, W.~Chen, W.~Ma, Q.~Ye, and T.-Y. Liu,
  ``\mbox{LightGBM}: A highly efficient gradient boosting decision tree,'' in
  \emph{Advances in neural information processing systems}, 2017, pp.
  3146--3154.

\bibitem{prokhorenkova2018catboost}
L.~Prokhorenkova, G.~Gusev, A.~Vorobev, A.~V. Dorogush, and A.~Gulin,
  ``\mbox{CatBoost}: unbiased boosting with categorical features,'' in
  \emph{Advances in neural information processing systems}, 2018, pp.
  6638--6648.

\bibitem{torres2019regression}
A.~Torres-Barr{\'a}n, {\'A}.~Alonso, and J.~R. Dorronsoro, ``Regression tree
  ensembles for wind energy and solar radiation prediction,''
  \emph{Neurocomputing}, vol. 326, pp. 151--160, 2019.

\bibitem{hocquenghem1959codes}
A.~Hocquenghem, ``Codes correcteurs d’erreurs,'' \emph{Chiffres}, vol.~2,
  no.~2, pp. 147--56, 1959.

\bibitem{bose1960class}
R.~C. Bose and D.~K. Ray-Chaudhuri, ``On a class of error correcting binary
  group codes,'' \emph{Information and control}, vol.~3, no.~1, pp. 68--79,
  1960.

\bibitem{Dietterich95}
T.~G. Dietterich and G.~Bakiri, ``Solving multiclass learning problems via
  error-correcting output codes,'' \emph{Journal of Artificial Intelligence
  Research}, vol.~2, pp. 263--286, 1995.

\bibitem{pujol2006discriminant}
O.~Pujol, P.~Radeva, and J.~Vitria, ``Discriminant \mbox{ECOC}: A heuristic
  method for application dependent design of error correcting output codes,''
  \emph{IEEE Transactions on Pattern Analysis and Machine Intelligence},
  vol.~28, no.~6, pp. 1007--1012, 2006.

\bibitem{bautista2012}
M.~{\'A}. Bautista, S.~Escalera, X.~Bar{\'o}, P.~Radeva, J.~Vitri{\`a}, and
  O.~Pujol, ``Minimal design of error-correcting output codes,'' \emph{Pattern
  Recognition Letters}, vol.~33, no.~6, pp. 693--702, 2012.

\bibitem{james1998error}
G.~James and T.~Hastie, ``The error coding method and picts,'' \emph{Journal of
  Computational and Graphical statistics}, vol.~7, no.~3, pp. 377--387, 1998.

\bibitem{james1998majority}
G.~James, ``Majority vote classifiers: theory and applications,'' Ph.D.
  dissertation, Stanford University, 1998.

\bibitem{EscaleraDecoding}
S.~Escalera, O.~Pujol, and P.~Radeva, ``On the decoding process in ternary
  error-correcting output codes,'' \emph{IEEE Trans. Pattern Anal. Mach.
  Intell.}, vol.~32, pp. 120--134, January 2010.

\bibitem{xiao2010speech}
L.~Xiao-Feng, Z.~Xue-ying, and D.~Ji-Kang, ``Speech recognition based on
  support vector machine and error correcting output codes,'' in \emph{2010
  First International Conference on Pervasive Computing, Signal Processing and
  Applications}.\hskip 1em plus 0.5em minus 0.4em\relax IEEE, 2010, pp.
  336--339.

\bibitem{ye2011pedestrian}
Q.~Ye, J.~Liang, and J.~Jiao, ``Pedestrian detection in video images via error
  correcting output code classification of manifold subclasses,'' \emph{IEEE
  Transactions on Intelligent Transportation Systems}, vol.~13, no.~1, pp.
  193--202, 2011.

\bibitem{smith2015facial}
R.~S. Smith and T.~Windeatt, ``Facial action unit recognition using multi-class
  classification,'' \emph{Neurocomputing}, vol. 150, pp. 440--448, 2015.

\bibitem{gu2019active}
S.~Gu, Y.~Cai, J.~Shan, and C.~Hou, ``Active learning with error-correcting
  output codes,'' \emph{Neurocomputing}, vol. 364, pp. 182--191, 2019.

\bibitem{zhang2020adversarial}
B.~Zhang, B.~Tondi, and M.~Barni, ``On the adversarial robustness of
  \mbox{DNNs} based on error correcting output codes,'' \emph{arXiv preprint
  arXiv:2003.11855}, 2020.

\bibitem{krizhevsky2009learning}
A.~Krizhevsky and G.~Hinton, ``Learning multiple layers of features from tiny
  images,'' Citeseer, Tech. Rep., 2009.

\bibitem{netzer2011reading}
Y.~Netzer, T.~Wang, A.~Coates, A.~Bissacco, B.~Wu, and A.~Y. Ng, ``Reading
  digits in natural images with unsupervised feature learning,'' 2011.

\bibitem{hughes2015open}
D.~Hughes, M.~Salath{\'e} \emph{et~al.}, ``An open access repository of images
  on plant health to enable the development of mobile disease diagnostics,''
  \emph{arXiv preprint arXiv:1511.08060}, 2015.

\bibitem{efron1992bootstrap}
B.~Efron, ``Bootstrap methods: another look at the jackknife,'' in
  \emph{Breakthroughs in statistics}.\hskip 1em plus 0.5em minus 0.4em\relax
  Springer, 1992, pp. 569--593.

\bibitem{schapire2003boosting}
R.~E. Schapire, ``The boosting approach to machine learning: An overview,'' in
  \emph{Nonlinear estimation and classification}.\hskip 1em plus 0.5em minus
  0.4em\relax Springer, 2003, pp. 149--171.

\bibitem{wolpert1992stacked}
D.~H. Wolpert, ``Stacked generalization,'' \emph{Neural networks}, vol.~5,
  no.~2, pp. 241--259, 1992.

\bibitem{lorena2010building}
A.~C. Lorena and A.~C. De~Carvalho, ``Building binary-tree-based multiclass
  classifiers using separability measures,'' \emph{Neurocomputing}, vol.~73,
  no. 16-18, pp. 2837--2845, 2010.

\bibitem{ethem99}
E.~Alpaydin and E.~Mayoraz, ``Learning error-correcting output codes from
  data,'' in \emph{Proceedings of the 9th International Conference on
  Artificial Neural Networks (ICANN 1999).}, vol.~2, 1999, pp. 743 --748.

\bibitem{flipiscis2010}
C.~Zor, B.~Yanikoglu, T.~Windeatt, and E.~Alpaydin, ``\mbox{FLIP-ECOC}: a
  greedy optimization of the \mbox{ECOC} matrix,'' in \emph{Proceedings of the
  25th International Symposium on Computer and Information Sciences (ISCIS
  2010)}.\hskip 1em plus 0.5em minus 0.4em\relax Springer, 2010, pp. 149 --
  154.

\bibitem{beamECOC2016}
C.~Zor, B.~A. Yanikoglu, E.~Merdivan, T.~Windeatt, J.~Kittler, and E.~Alpaydin,
  ``\mbox{BeamECOC}: {A} local search for the optimization of the \mbox{ECOC}
  matrix,'' in \emph{23rd International Conference on Pattern Recognition,
  {ICPR} 2016, Canc{\'{u}}n, Mexico, December 4-8, 2016}, 2016, pp. 198--203.

\bibitem{aly2018multi}
S.~A. Aly and B.~Yanikoglu, ``Multi-label networks for face attributes
  classification,'' in \emph{2018 IEEE International Conference on Multimedia
  \& Expo Workshops (ICMEW)}.\hskip 1em plus 0.5em minus 0.4em\relax IEEE,
  2018, pp. 1--6.

\bibitem{szegedy2016rethinking}
C.~Szegedy, V.~Vanhoucke, S.~Ioffe, J.~Shlens, and Z.~Wojna, ``Rethinking the
  inception architecture for computer vision,'' in \emph{Proceedings of the
  IEEE conference on computer vision and pattern recognition}, 2016, pp.
  2818--2826.

\bibitem{chollet2017xception}
F.~Chollet, ``Xception: Deep learning with depthwise separable convolutions,''
  in \emph{Proceedings of the IEEE conference on Computer Vision and Pattern
  Recognition}, 2017, pp. 1251--1258.

\bibitem{hu2018squeeze}
J.~Hu, L.~Shen, and G.~Sun, ``Squeeze-and-excitation networks,'' in
  \emph{Proceedings of the IEEE conference on computer vision and pattern
  recognition}, 2018, pp. 7132--7141.

\bibitem{deng2009imagenet}
J.~Deng, W.~Dong, R.~Socher, L.-J. Li, K.~Li, and L.~Fei-Fei, ``Imagenet: A
  large-scale hierarchical image database,'' in \emph{2009 IEEE conference on
  computer vision and pattern recognition}.\hskip 1em plus 0.5em minus
  0.4em\relax Ieee, 2009, pp. 248--255.

\bibitem{vedaldi2015matconvnet}
A.~Vedaldi and K.~Lenc, ``Matconvnet: Convolutional neural networks for
  matlab,'' in \emph{Proceedings of the 23rd ACM international conference on
  Multimedia}, 2015, pp. 689--692.

\bibitem{kuralenok2019factorized}
I.~E. Kuralenok, Y.~Rebryk, R.~Solovev, and A.~Ermilov, ``Factorized multiclass
  boosting,'' \emph{arXiv preprint arXiv:1909.04904}, 2019.

\bibitem{mohanty2016using}
S.~P. Mohanty, D.~P. Hughes, and M.~Salath{\'e}, ``Using deep learning for
  image-based plant disease detection,'' \emph{Frontiers in plant science},
  vol.~7, p. 1419, 2016.

\bibitem{too2019comparative}
E.~C. Too, L.~Yujian, S.~Njuki, and L.~Yingchun, ``A comparative study of
  fine-tuning deep learning models for plant disease identification,''
  \emph{Computers and Electronics in Agriculture}, vol. 161, pp. 272--279,
  2019.

\bibitem{kamal2019depthwise}
K.~Kamal, Z.~Yin, M.~Wu, and Z.~Wu, ``Depthwise separable convolution
  architectures for plant disease classification,'' \emph{Computers and
  Electronics in Agriculture}, vol. 165, p. 104948, 2019.

\end{thebibliography}

\newpage
\begin{IEEEbiography}[{\includegraphics[width=1in,height=1.25in,clip,keepaspectratio]{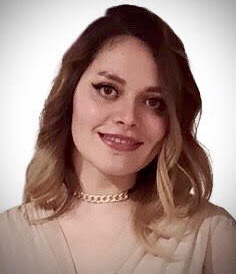}}]{Sara Atito Ali Ahmed} received her Bsc. in computer science from Ain Shams University, Egypt, in 2011. Her Msc. degree was a collaboration between Nile University, Egypt and TU Berlin, Germany in 2014, working on vehicles detection and tracking in  crowded scenes and sensitivity analysis of deep neural networks. In 2013, she was an intern in Speech \& Sound Group in Sony Deutschland GmbH - Stuttgart, Germany, working on character recognition in natural images. 
She is a PhD student at Sabanci University, with a thesis on deep learning ensembles  for image understanding. Currently, she is a fellow researcher in CVSSP group in University of Surrey, UK working on detection and generation of face morphing. 
\end{IEEEbiography}

\begin{IEEEbiography}[{\includegraphics[width=1in,height=1.25in,clip,keepaspectratio]{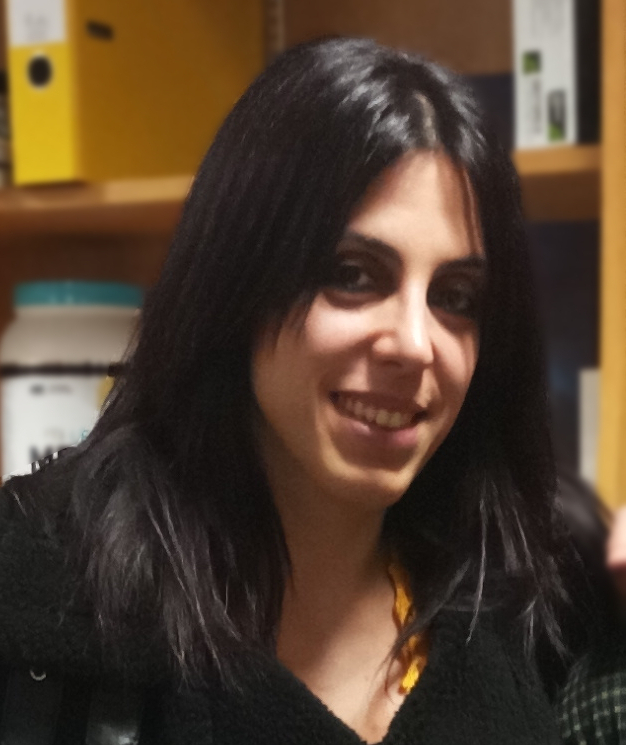}}]{Cemre Zor} is a senior research associate at the Centre for Medical Image Computing at University College London. She was previously a research associate in the Centre for Vision, Speech and Signal Processing / University of Surrey, where she also obtained her M.Sc. and Ph.D. degrees. Her current research interests include multi-modal predictive analysis and anomaly detection in neuroscience data, multiple classifier systems and the theory of classification.
\end{IEEEbiography}

\begin{IEEEbiography}[{\includegraphics[width=1in,height=1.25in,clip]{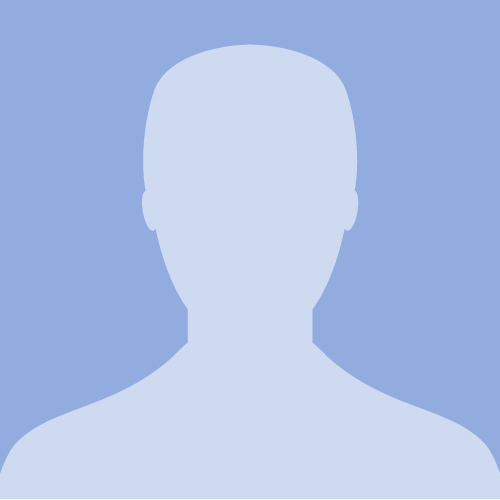}}]{Muhammad Awais}
received the B.Sc. degree in Mathematics and Physics from the AJK University in 2001, B.Sc. degree in computer engineering from UET Taxila in 2005, M.Sc in signal processing and machine intelligence and PhD in machine learning from the University of Surrey in 2008 and 2011.
He is currently a senior research fellow at the Centre for Vision, Speech and Signal Processing (CVSSP) at the University of Surrey.
His research interests include machine learning, deep learning, self(un,semi)-supervised learning, NLP, audio-visual analysis, medical image analysis and computer vision.
\end{IEEEbiography}

\begin{IEEEbiography}[{\includegraphics[width=1in,height=1.25in,clip,keepaspectratio]{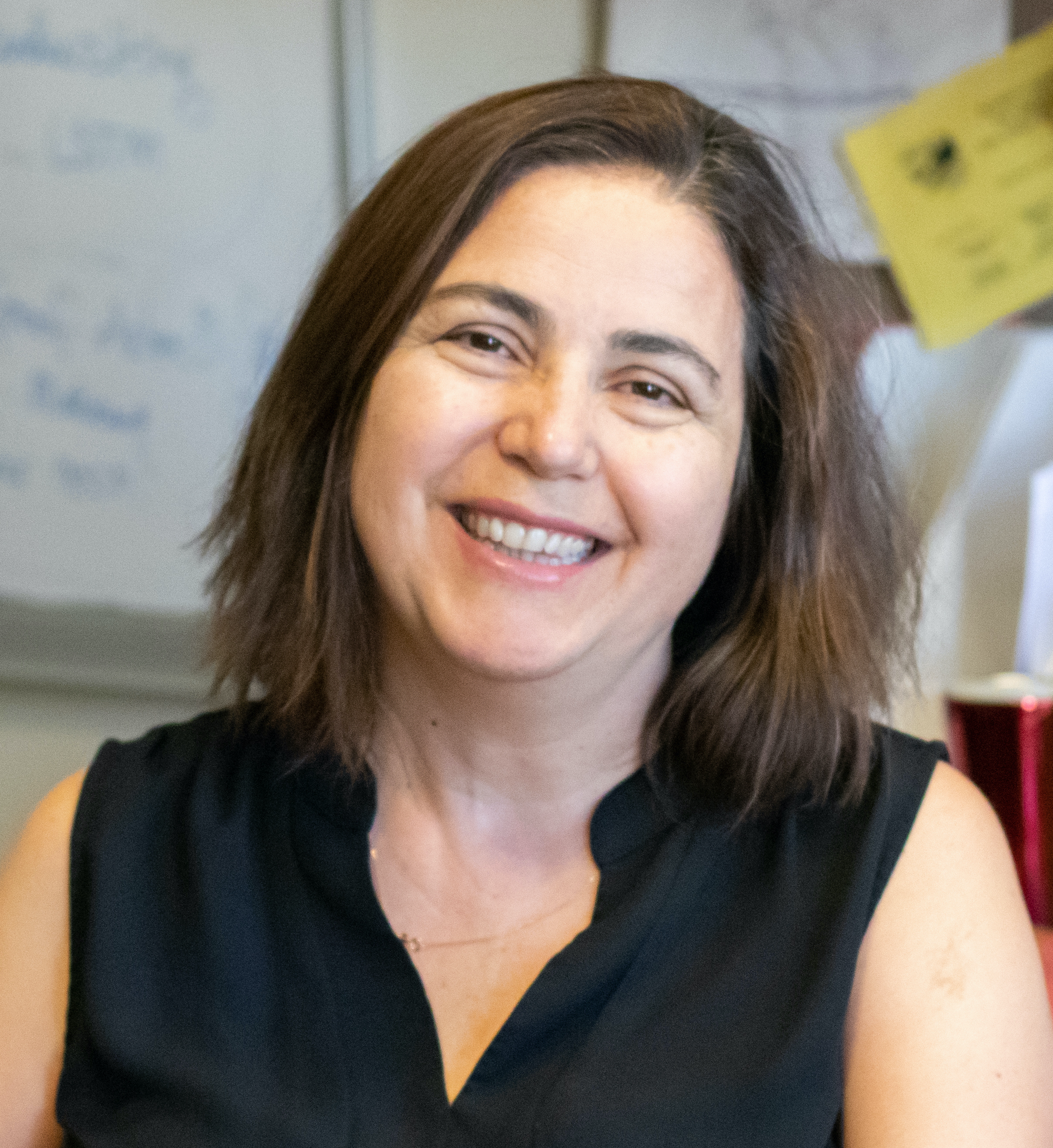}}]{Berrin Yanikoglu} received a double major in Computer Science and Mathematics from Bogazici University, Turkey in 1988 and her Ph.D. degreE in Computer Science from Dartmouth College, USA in 1993. She is a Professor of Computer Science and Director of the Center of Excellence in Data Analytics (VERIM), at Sabanci University, Istanbul, Turkey. Prof. Yanikoglu worked at Rockefeller University, Xerox Imaging Systems and IBM Almaden Research Center, before joining Sabanci University in 2000. Her research focuses on image/video understanding, biometric verification and privacy, and handwriting recognition. She is an Editor for the Turkish Journal of Electrical Engineering and Computer Science. 
\end{IEEEbiography}

\begin{IEEEbiography}[{\includegraphics[width=1in,height=1.25in,clip,keepaspectratio]{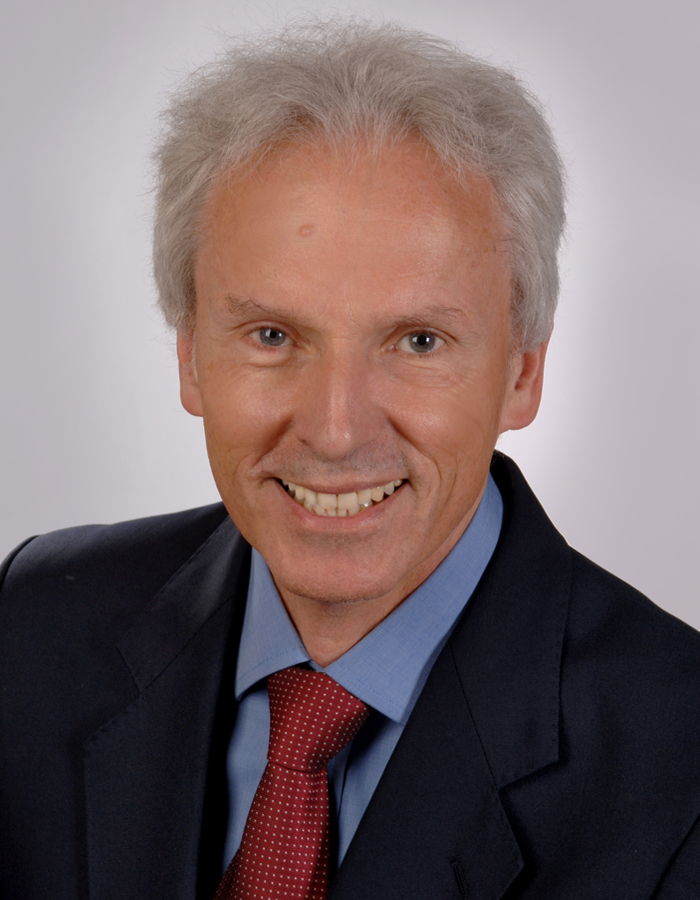}}]{Josef Kittler}
(M'74-LM'12) received the B.A., Ph.D., and D.Sc. degrees from the University of Cambridge, in 1971, 1974, and 1991, respectively.
He is a distinguished Professor of Machine Intelligence at the Centre for Vision, Speech and Signal Processing, University of Surrey, Guildford, U.K.
He conducts research in biometrics, video and image dataset retrieval, medical image analysis, and
cognitive vision. He published the textbook Pattern Recognition: A Statistical Approach and over 700 scientific papers. 
His publications have been cited more than 68,000 times (Google Scholar).

He is series editor of Springer Lecture Notes on Computer Science. He currently serves on the Editorial Boards of Pattern Recognition Letters, Pattern Recognition and Artificial Intelligence, Pattern Analysis and Applications. He also served as a member of the Editorial Board of IEEE Transactions on Pattern Analysis and Machine Intelligence during 1982-1985. He served on the Governing Board of the International Association for Pattern Recognition (IAPR) as one of the two British representatives during the period 1982-2005, President of the IAPR during 1994-1996. Currently he is a member of the KS Fu Prize Committee of IAPR.
\end{IEEEbiography}

\end{document}